\documentclass[10pt,twocolumn,letterpaper]{article}
\usepackage[pagenumbers]{cvpr}
\usepackage{amsthm}
\usepackage{algorithm}
\usepackage{algpseudocode}
\usepackage{multirow}
\usepackage{tabularx}
\usepackage{placeins}
\usepackage{colortbl}
\usepackage{pifont}

\definecolor{proudtableheadbg}{RGB}{239,241,243}
\definecolor{proudtableshade}{RGB}{245,246,247}
\definecolor{proudtableresult}{rgb}{0.88,1,1}
\definecolor{proudtablerule}{RGB}{90,95,100}
\newcommand{\proudtabsetup}{%
  \small
  \setlength{\tabcolsep}{4pt}%
  \renewcommand{\arraystretch}{1.04}%
  \setlength{\extrarowheight}{1.5pt}%
  \setlength{\arrayrulewidth}{0.3pt}%
  \setlength{\heavyrulewidth}{0.6pt}%
  \setlength{\lightrulewidth}{0.3pt}%
  \setlength{\cmidrulewidth}{0.3pt}%
  \setlength{\aboverulesep}{0pt}%
  \setlength{\belowrulesep}{0pt}%
  \arrayrulecolor{proudtablerule}%
  \captionsetup{justification=raggedright,singlelinecheck=false,skip=5pt}%
}
\newcommand{\prouddensetable}{\fontsize{8.5}{10.2}\selectfont}
\newcommand{\proudmaintabsetup}{%
  \proudtabsetup
  \renewcommand{\arraystretch}{1}%
  \setlength{\extrarowheight}{0pt}%
}
\newcommand{\proudanalysistabsetup}{%
  \proudmaintabsetup
  \prouddensetable
  \renewcommand{\arraystretch}{1.12}%
  \setlength{\tabcolsep}{2pt}%
  \setlength{\aboverulesep}{0.4ex}%
  \setlength{\belowrulesep}{0.65ex}%
  \setlength{\heavyrulewidth}{0.65pt}%
  \setlength{\lightrulewidth}{0.35pt}%
}
\newcommand{\proudcmark}{\ding{51}}
\newcommand{\proudxmark}{\ding{55}}
\DeclareMathSizes{8.5}{8}{6}{5}
\newcolumntype{C}[1]{>{\hsize=#1\hsize\linewidth=\hsize\centering\arraybackslash}X}
\newcommand{\proudtablehead}[1]{{\bfseries\shortstack[c]{#1}}}

\newtheorem{proposition}{Proposition}

\definecolor{cvprblue}{rgb}{0.21,0.49,0.74}
\usepackage[pagebackref,breaklinks,colorlinks,allcolors=cvprblue]{hyperref}

\def\confName{CVPR}
\def\confYear{2027}

\title{Projector Subspace Optimization for Multimodal Continual Instruction Tuning}
\author{
\textbf{Hao Yu}$^{1,2}$, \quad \textbf{Quan Cheng}$^{1,2}$, \quad \textbf{Lijun Zhang}$^{1,2,}$\thanks{Lijun Zhang is the corresponding author.}\\
{\normalsize $^1$School of Artificial Intelligence, Nanjing University}\\
{\normalsize $^2$State Key Laboratory for Novel Software Technology, Nanjing University}\\
{\small\ttfamily \{yuhao, chengq, zhanglj\}@lamda.nju.edu.cn}
}

\date{}
\hypersetup{
  pdftitle={Projector Subspace Optimization for Multimodal Continual Instruction Tuning},
  pdfauthor={Hao Yu, Quan Cheng, Lijun Zhang}
}

\usepackage{microtype}

\begin{document}
\raggedbottom
\maketitle

\begin{abstract}
Multimodal Large Language Models (MLLMs) achieve strong performance through instruction tuning, yet adapting them to sequential tasks requires Multimodal Continual Instruction Tuning (MCIT). Existing replay-free methods expand prompts or routed LoRA experts across multi-layer decoders, incurring heavy parameter and routing overhead while shifting internal feature distributions. Instead, we identify the final linear layer of the vision--language projector as a high-capacity, coordinate-stationary bridge, where sequential updates easily induce interface drift by displacing historical visual tokens. While low-rank subspace adaptation offers a compact remedy, selecting directions solely from historical statistics or task gradients sacrifices plasticity or stability, and unconstrained multi-step updates still accumulate large drift. To resolve both issues, we propose Projector Refinement through Optimized Update Directions (PROUD), a continual subspace optimization framework for the terminal projector layer. At task onset, PROUD combines retained input statistics with initial gradients to analytically extract a metric-orthonormal subspace maximizing first-order task gain under a drift budget. During training, it regularizes updates within this fixed subspace via an exact Frobenius penalty before losslessly merging them with zero inference or routing overhead. Experiments on UCIT and CoIN across MLLM backbones show that PROUD outperforms recent replay-free baselines in continual accuracy while curbing forgetting with a fraction of trainable parameters.
\end{abstract}

\begin{figure}[t]
  \centering
  \includegraphics[width=\linewidth]{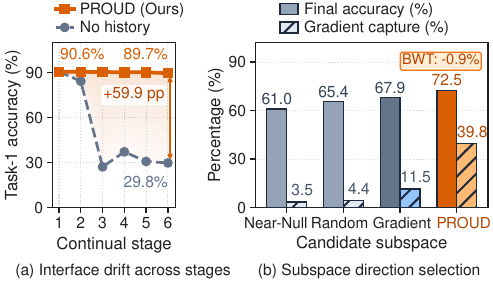}
  \caption{\textbf{Diagnosing interface drift and subspace selection on UCIT.} (a) Task-1 retention under PROUD and its variant without historical statistics, with matched Stage-1 starts. (b) Held-out whitened gradient capture versus final accuracy and backward transfer (BWT) across rank-128 subspaces. PROUD combines high whitened gradient capture with minimal forgetting.}
  \label{fig:teaser}
\end{figure}

\begingroup
\AtBeginEnvironment{itemize}{\catcode`\&=12\relax}
\section{Introduction}
\label{sec:intro}

Multimodal Large Language Models (MLLMs) have demonstrated strong capabilities across diverse vision--language tasks through instruction tuning~\cite{dai2023instructblip,liu2023llava,liu2024improved}. However, real-world environments are dynamic, requiring deployed models to continually master sequential tasks with shifting visual domains and instruction styles, a setting known as Multimodal Continual Instruction Tuning (MCIT)~\cite{chen2024coin,he2023continual}. A central challenge in MCIT is catastrophic forgetting, where adapting to incoming tasks overwrites previously acquired multimodal alignments and degrades performance on earlier tasks~\cite{guo2025hide,mccloskey1989catastrophic,wang2024comprehensive,zhou2024continual,zhai2024investigating}.

To mitigate forgetting without replaying historical data, recent MCIT methods typically freeze the pre-trained backbones and insert task-specific prompts or Low-Rank Adaptation (LoRA) experts across multi-layer language decoders, often relying on dynamic routing at inference~\cite{chen2024coin,guo2025hide,yu2025progressive}. However, expanding per-task modules and routers across dozens of decoder layers incurs substantial parameter overhead and routing latency, while updating multi-layer weights continually shifts internal feature distributions and invalidates historical statistics. In contrast, the vision--language projector serves as the core cross-modal bridge translating visual features into language token embeddings. Notably, we find that adapting only the final linear layer of the projector closely matches the joint-training accuracy of the full projector and rivals multi-layer decoder tuning, while keeping upstream features strictly stationary across tasks. Yet because all tasks share this compact bridge, updating it sequentially without protection induces severe \emph{interface drift}: the projected visual tokens of earlier tasks shift over time, distorting fine-grained spatial coordinates for visual grounding and erasing localized cues for scientific or OCR reasoning. As shown in Figure~\ref{fig:teaser}(a), adapting without historical protection severely degrades earlier tasks across stages, whereas our approach maintains stable long-term~retention.

\begingroup
\clubpenalty=10000
Mitigating this interface drift under a compact low-rank budget requires examining how weight updates interact with both historical features and incoming task gradients. As revealed by our diagnostic analysis in Figure~\ref{fig:teaser}(b), existing single-criterion subspaces~\cite{liang2024inflora,saha2021gpm,wang2024loraga,wang2023olora,corda2024} suffer from a fundamental \emph{directional mismatch} at the projector interface. Confining updates solely to low-energy historical directions protects prior tokens, yet captures minimal task gradient energy and severely restricts plasticity. Conversely, aligning updates directly with dominant task gradients captures immediate learning signals, but collides with high-energy historical directions and exacerbates forgetting. Moreover, static subspace selection alone is insufficient: across shifting upstream layers, stored historical statistics quickly become obsolete without replay, while even at a stationary interface, unconstrained multi-step optimization within a chosen subspace still accumulates large weight updates that displace prior visual tokens over an epoch. Consequently, mitigating interface drift requires a unified design that preserves exact historical statistics at a stationary interface, balances task utility against historical interference when selecting update directions, and controls cumulative update magnitudes throughout multi-step training.\par
\endgroup

To this end, we propose \underline{P}rojector \underline{R}efinement through \underline{O}ptimized \underline{U}pdate \underline{D}irections (PROUD), a continual subspace optimization framework for replay-free MCIT. Instead of inserting expandable modules across shifting decoder layers, PROUD adapts only the terminal projector layer, leveraging frozen upstream representations to maintain a stationary coordinate system to track historical token drift. At task onset, PROUD combines retained input statistics with initial gradients to analytically identify a low-rank subspace maximizing first-order task improvement under a historical drift budget. During training, it restricts updates to this metric-orthonormal subspace, reducing in-stage drift control to an exact Frobenius penalty that curbs multi-stage token drift before losslessly merging the update into the projector with zero inference parameter growth or routing~overhead.

In summary, our main contributions are as follows:
\begin{itemize}
    \item We identify \emph{interface drift} at the vision--language projector as a key bottleneck in fixed-capacity MCIT, and establish the terminal projector layer as a high-capacity, coordinate-stationary bridge for continual multimodal alignment across sequential tasks.
    \item We propose PROUD, a continual subspace optimization framework that couples closed-form benefit-to-cost subspace selection at task onset with exact isometric drift regularization during training and lossless post-stage weight~merging.
    \par
    \begingroup
    \interlinepenalty=10000
    \item Extensive experiments on UCIT and CoIN across multiple MLLM backbones demonstrate that PROUD achieves superior continual performance over competing replay-free methods, using a fraction of the trainable~parameters.\par
    \endgroup
\end{itemize}
\endgroup

\section{Related work}
\label{sec:related_work}

\noindent\textbf{Multimodal continual instruction tuning.}
Continual instruction tuning aims to equip Multimodal Large Language Models (MLLMs) with new task capabilities sequentially while mitigating catastrophic forgetting~\cite{chen2024coin,he2023continual,zhai2024investigating}.
While classical continual learning relies on parameter regularization~\cite{kirkpatrick2017overcoming} or feature projection~\cite{zeng2019owm} in models trained from scratch,
recent MLLM approaches freeze the pre-trained backbones and update only lightweight parameter-efficient modules~\cite{hu2022lora,wang2024comprehensive,zhou2024continual}.
Prompt-based methods maintain learnable prompt pools or dual-modality prompts, retrieving or combining them for each input during inference, as in L2P~\cite{wang2022learning}, DualPrompt~\cite{wang2022dualprompt}, CODA-Prompt~\cite{smith2023coda}, and ModalPrompt~\cite{zeng2025modalprompt}.
Complementary LoRA-based methods allocate task-specific low-rank experts and combine them via dynamic routing or hierarchical fusion, including Continual LLaVA~\cite{cao2024continual}, MoELoRA~\cite{chen2024coin}, CL-MoE~\cite{huai2025cl}, ProgLoRA~\cite{yu2025progressive}, and HiDe-LLaVA~\cite{guo2025hide}, whereas SEFE~\cite{chen2025sefe} regularizes top-magnitude LoRA updates alongside answer-style diversification.
Despite their effectiveness, these approaches either expand per-task module pools with inference-time routing overhead or continually modify multi-layer decoder representations across sequential tasks.

\smallskip
\noindent\textbf{Subspace projection and low-rank adaptation.}
To control forgetting without replay or parameter growth, subspace-based continual learning restricts parameter updates using historical information. Building on Gradient Projection Memory (GPM)~\cite{saha2021gpm},
recent adapter methods constrain updates to subspaces orthogonal to prior adapter weights as in O-LoRA~\cite{wang2023olora}, or to the low-energy complement of historical input features as in InfLoRA~\cite{liang2024inflora}.
TRGP~\cite{lin2022trgp} relaxes rigid orthogonality by selecting correlated prior subspaces and learning per-task scaling matrices within a trust region.
Meanwhile, recent single-task LoRA variants improve memory efficiency or convergence speed: LoRA-FA~\cite{zhang2023lorafa} freezes a random input projection, LoRA-GA~\cite{wang2024loraga} aligns initial adapters with dominant task gradients, and CorDA~\cite{corda2024} decomposes weight--activation products.
However, continual null-space methods rely solely on historical representations when defining the subspace, whereas gradient-aligned adapters neglect cross-task interference. Our work bridges this gap by jointly optimizing task utility and historical drift at a stationary projector interface.

\section{Preliminaries}
\label{sec:preliminaries}

We first formalize the replay-free MCIT setting and then analyze the stability--plasticity trade-off in low-rank adaptation.

\subsection{Multimodal continual instruction tuning}
\label{sec:mcit_protocol}

Multimodal continual instruction tuning aims to sequentially adapt a pre-trained MLLM across shifting task distributions without forgetting prior capabilities. Formally, we consider an MLLM consisting of a vision encoder, a vision--language projector, and a language decoder~\cite{liu2023llava,liu2024improved,li2023blip2}. Given an input image $I$ and a text instruction $q$, the vision encoder and projector map the image into visual tokens, which are concatenated with the instruction embeddings and fed into the decoder to generate a response $y = (y_1, \dots, y_L)$.

In Multimodal Continual Instruction Tuning (MCIT), the model encounters a sequence of $T$ tasks $\{\mathcal{D}_1, \dots, \mathcal{D}_T\}$, where $\mathcal{D}_t = \{(I_i, q_i, y_i)\}_{i=1}^{N_t}$ denotes the training set of stage $t$. Under the replay-free protocol, historical data $\mathcal{D}_{<t}$ and test-time task identities are inaccessible. We additionally target a fixed-capacity deployment without inference routing. At stage $t$, optimizing the adaptable parameters $W$ solely on $\mathcal{D}_t$ minimizes the stage cross-entropy loss:
\begin{equation}
    \mathcal{L}_t^{\mathrm{CE}}(W) = -\sum_{i=1}^{N_t}\sum_{k=1}^{L_i}\frac{\log p_W(y_{i,k} \mid I_i, q_i, y_{i,<k})}{N_t L_i}.
    \label{eq:ce_loss}
\end{equation}
Since $\mathcal{D}_{<t}$ is unavailable, unconstrained optimization of Eq.~\eqref{eq:ce_loss} overwrites previously acquired alignments and causes catastrophic forgetting across tasks.

\subsection{Low-rank subspace adaptation}
\label{sec:subspace_prelim}

Parameter-efficient continual learning commonly restricts weight updates to low-rank subspaces to reduce training cost and cross-task interference. To adapt a linear layer $W \in \mathbb{R}^{m\times d}$ at stage $t$, low-rank methods constrain the update relative to $W_{t-1}$ within a rank-$r$ subspace ($r \ll \min(m,d)$):
\begin{equation}
    \Delta W_t = W - W_{t-1} = B_t P_t^\top,
    \label{eq:low_rank_update}
\end{equation}
where the columns of $P_t \in \mathbb{R}^{d\times r}$ span the $r$-dimensional input subspace and $B_t \in \mathbb{R}^{m\times r}$ is the outgoing factor. Existing rehearsal-free methods select $P_t$ using a single criterion. Null-space projection methods~\cite{saha2021gpm,liang2024inflora} choose $P_t$ from low-energy directions of historical features $h \in \mathbb{R}^d$ so that $\Delta W_t h \approx \mathbf{0}$, whereas gradient-aligned initialization~\cite{wang2024loraga} selects $P_t$ from the top right singular vectors of the initial gradient $\bar{G}_t = \nabla_W \mathcal{L}_t^{\mathrm{CE}}(W_{t-1}; \mathcal{B}_t)$ on a preview subset $\mathcal{B}_t \subset \mathcal{D}_t$ to maximize $-\langle \bar{G}_t, \Delta W_t \rangle_F$.

To expose the tension between plasticity and stability, let $h = \phi(I) \in \mathbb{R}^d$ denote an upstream visual-token feature. Substituting $\Delta W_t = B_t P_t^\top$ into the first-order task loss reduction and expected historical output perturbation $\|\Delta W_t h\|_2^2$ reveals a coupled trade-off \mbox{across $\phi(I)$, $P_t$, and $B_t$:}
\begin{equation}
    \underset{\text{Plasticity benefit on }\mathcal{D}_t}{-\langle \bar{G}_t P_t, B_t \rangle_F}
    \;\text{vs.}\;
    \underset{\text{Historical interface drift on }\mathcal{D}_{<t}}{\mathrm{tr}\!\left(B_t P_t^\top \mathbb{E}\!\big[hh^\top\big] P_t B_t^\top\right)}.
    \label{eq:tradeoff}
\end{equation}

Equation~\eqref{eq:tradeoff} exposes three coupled obstacles in continual low-rank adaptation. First, regarding \emph{feature non-stationarity}, updating backbones or upstream adapters continually shifts the input mapping $\phi(I)$ across stages, invalidating accumulated statistics $\mathbb{E}[hh^\top]$ without replay. Second, regarding \emph{onset directional mismatch}, subspace selection must balance projected task gradients $\bar{G}_t P_t$ against historical interference $P_t^\top \mathbb{E}[hh^\top] P_t$. Third, regarding \emph{in-stage magnitude drift}, unconstrained multi-step optimization amplifies $\|B_t\|_F$ and inflates drift even under a well-chosen $P_t$, motivating a unified framework.

\begin{figure*}[t]
  \centering
  \includegraphics[width=0.95\textwidth]{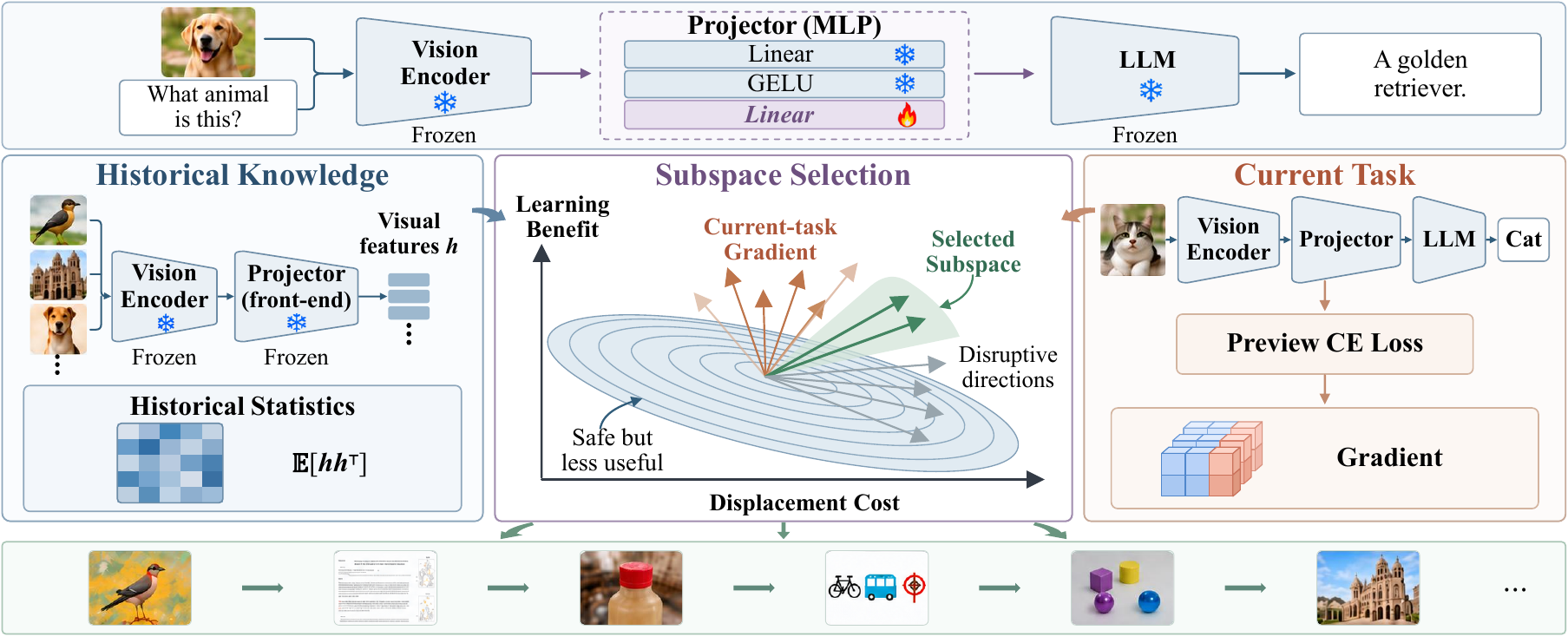}
  \caption{\textbf{Overview of PROUD.} Frozen upstream layers keep the final projector's input features stationary. At each task, historical statistics and the initial task gradient select a metric-orthonormal input basis. The basis stays fixed while the outgoing factor is trained; its update is merged before the next task.}
  \label{fig:framework}
\end{figure*}

\section{Methodology}
\label{sec:methodology}

As illustrated in Figure~\ref{fig:framework}, PROUD resolves the three obstacles in Eq.~\eqref{eq:tradeoff} through a unified three-stage framework. First, we restrict adaptation to the terminal projector layer to preserve stationary input coordinates and maintain an exact stage-balanced historical metric. Second, at task onset, we couple this metric with initial task gradients to analytically extract the optimal low-rank update basis. Finally, we enforce isometric in-stage regularization within this fixed basis before merging updates into the shared projector.

\subsection{Stationary projector interface}
\label{sec:stationary_metric}

\noindent\textbf{Coordinate stationarity.} To preserve exact historical feature statistics across tasks without data replay, we restrict adaptation to the final linear layer of the vision--language projector while keeping the vision encoder, preceding projector layers, and language decoder frozen. For the $j$-th visual token of an image $I$, let $h_j = \phi(I)_j \in \mathbb{R}^d$ denote the input feature to the final projector layer, where $\phi(\cdot)$ represents the frozen vision encoder and upstream projector layers. The final projector layer maps $h_j$ into $\mathbb{R}^m$:
\begin{equation}
    z_j = W h_j + b, \qquad W \in \mathbb{R}^{m\times d}, \quad b \in \mathbb{R}^m.
    \label{eq:projector_linear}
\end{equation}
\looseness=-1 During continual tuning, we update only $W$ and keep $b$ frozen. Because the upstream mapping $\phi(\cdot)$ is fixed, any image $I$ produces the exact same input representation $h_j = \phi(I)_j$ across all stages $t \in \{1, \dots, T\}$, a property we refer to as \emph{coordinate stationarity}. Unlike earlier projector layers whose updates pass through intermediate nonlinearities before reaching the decoder, the terminal linear layer maps stationary features directly into language token embeddings.

\noindent\textbf{Stage-balanced historical metric.} To quantify historical drift without storing past images, we track token displacement under coordinate stationarity. Updating weights at stage $t$ from $W_{t-1}$ to $W = W_{t-1} + \Delta W$ alters the projector output on any historical token $h \in \mathbb{R}^d$ by $\Delta z = \Delta W h$, inducing a squared output displacement of $\|\Delta W h\|_2^2 = \operatorname{tr}(\Delta W h h^\top \Delta W^\top)$. Specifically, after each stage $s$, we compute the empirical second-moment matrix over $n_s$ sampled training images:
\begin{equation}
    \widehat{C}_s = \frac{1}{n_s}\sum_{i=1}^{n_s}\frac{1}{|\mathcal{J}_i|}\sum_{j\in\mathcal{J}_i} h_{i,j} h_{i,j}^\top \in \mathbb{R}^{d\times d},
    \label{eq:stage_cov}
\end{equation}
where $\mathcal{J}_i$ is a sampled subset of visual token indices from image $I_i$. Averaging within each image before cross-sample averaging prevents images with larger token counts from dominating $\widehat{C}_s$. Across stages, we initialize $C_{<1} = \mathbf{0}$ and incrementally maintain the stage-balanced average $C_{<t} = \frac{1}{t-1}\sum_{s<t}\widehat{C}_s$ for $t > 1$, with isotropic damping:
\begin{equation}
    M_t = C_{<t} + \rho I_d \succ 0,
    \label{eq:damped_metric}
\end{equation}
where $\rho > 0$ is a damping constant scaled by the initial token energy to ensure strict positive definiteness and anchor Stage-1 updates to the pre-trained projector $W_0$. The stage-wise displacement cost $q_t(\Delta W) \triangleq \operatorname{tr}(\Delta W M_t \Delta W^\top)$ then measures expected historical output perturbation:
\begin{equation}
    q_t(\Delta W) = \mathbb{E}_{h \sim \widehat{\mathcal{H}}_{<t}}\big[\|\Delta W h\|_2^2\big] + \rho \|\Delta W\|_F^2,
    \label{eq:drift_decomposition}
\end{equation}
where $\widehat{\mathcal{H}}_{<t}$ is the empirical historical token distribution. Because $\phi(\cdot)$ is frozen, $C_{<t}$ remains in a fixed coordinate system with a task-independent $\mathcal{O}(d^2)$ memory footprint, while gradients from Eq.~\eqref{eq:ce_loss} backpropagate through the frozen decoder to provide task supervision.

\subsection{Budgeted subspace selection}
\label{sec:subspace_selection}

At the start of stage $t$, we seek a rank-$r$ input subspace $P_t \in \mathbb{R}^{d\times r}$ that maximizes first-order task improvement on $\mathcal{D}_t$ within a historical displacement budget under $M_t$. Before training on $\mathcal{D}_t$, we sample a small preview subset $\mathcal{B}_t \subset \mathcal{D}_t$ and perform one forward and backward pass at fixed $W_{t-1}$ to obtain the initial task gradient $\bar{G}_t = \nabla_W \mathcal{L}_t^{\mathrm{CE}}(W_{t-1}; \mathcal{B}_t) \in \mathbb{R}^{m\times d}$. Under a first-order Taylor approximation, a candidate update $\Delta W$ reduces current-task loss by $-\langle \bar{G}_t, \Delta W \rangle_F$. Given a drift budget $\varepsilon^2 > 0$ under $q_t(\Delta W)$, we formulate subspace discovery as:
\begin{equation}
    \begin{aligned}
    \max_{\Delta W \in \mathbb{R}^{m\times d}}\quad &-\langle \bar{G}_t, \Delta W \rangle_F \\
    \text{s.t.}\quad &\operatorname{rank}(\Delta W) \le r, \quad q_t(\Delta W) \le \varepsilon^2.
    \end{aligned}
    \label{eq:constrained_opt}
\end{equation}

Factoring $\Delta W = B P^\top$ with $P^\top M_t P = I_r$ reduces $q_t(\Delta W) \le \varepsilon^2$ to $\|B\|_F \le \varepsilon$. Maximizing over $B$ by Cauchy--Schwarz gives $\varepsilon\|\bar{G}_t P\|_F$, so the optimal basis $P$ maximizes $\|\bar{G}_t P\|_F^2$ independently of any $\varepsilon > 0$. We then use Cholesky whitening $M_t = L_t L_t^\top$, where $L_t \in \mathbb{R}^{d\times d}$ is lower triangular. Let $K_t = \bar{G}_t L_t^{-\top} \in \mathbb{R}^{m\times d}$ denote the metric-whitened task gradient, and let $K_t = U_t \Sigma_t V_t^\top$ be its SVD with singular values $\sigma_1(K_t) \ge \dots \ge \sigma_{\min(m,d)}(K_t) \ge 0$. Selecting the top-$r$ right singular vectors $V_{t,r} \in \mathbb{R}^{d\times r}$ of $K_t$ captures the steepest descent directions in whitened space, while mapping them back via $L_t^{-\top}$ naturally preconditions the basis against \mbox{historically dominant directions}.

\begin{proposition}[Optimal Update Subspace]
\label{prop:optimal_subspace}
The optimal rank-$r$ input basis for Problem~\eqref{eq:constrained_opt} is given by
\begin{equation}
    P_t = L_t^{-\top} V_{t,r} \in \mathbb{R}^{d\times r}, \qquad P_t^\top M_t P_t = I_r,
    \label{eq:optimal_basis}
\end{equation}
which maximizes the metric-normalized gradient energy (proof in Appendix~\ref{app:optimal_subspace_proof}):
\begin{equation*}
    \max_{P^\top M_t P = I_r} \|\bar{G}_t P\|_F^2 = \sum_{i=1}^r \sigma_i^2(K_t).
\end{equation*}
\end{proposition}

\noindent\textbf{Discussion.} Proposition~\ref{prop:optimal_subspace} quantifies the directional benefit-to-cost ratio via $\sigma_i^2(K_t)$ and unifies prior single-criterion paradigms as limiting cases: with no historical statistics ($C_{<t} = \mathbf{0}$), $P_t$ spans the top-$r$ right singular vectors of $\bar{G}_t$ for pure gradient alignment~\cite{wang2024loraga}; under isotropic task gradients ($\bar{G}_t^\top \bar{G}_t \propto I_d$), it selects the minor eigenvectors of $C_{<t}$ for historical near-null projection~\cite{liang2024inflora,saha2021gpm}. Joint ranking avoids both extremes by filtering out low-gradient null directions and high-energy historical directions.

\setcounter{dbltopnumber}{2}
\renewcommand{\dbltopfraction}{0.8}
\begin{table*}[t]
\centering
\proudmaintabsetup
\prouddensetable
\setlength{\tabcolsep}{3pt}
\caption{Main results on the UCIT benchmark with LLaVA-v1.5-7B as the backbone. Params denotes the number of parameters updated per task (merged after each task for PROUD). The best and second-best results are highlighted in \textbf{bold} and \underline{underline}, respectively.}
\label{tab:ucit_benchmark}
\begin{tabularx}{\linewidth}{lc*{6}{C{1}}*{3}{C{1}}}
\toprule
\rowcolor{proudtableheadbg}
\textbf{Method} & \textbf{Params} & \textbf{ImgNet-R} & \textbf{ArxivQA} & \textbf{VizWiz} & \textbf{IconQA} & \textbf{CLEVR} & \textbf{Flickr30k} & \proudtablehead{Last $\uparrow$} & \proudtablehead{Avg $\uparrow$} & \proudtablehead{BWT $\uparrow$} \\
\midrule
Zero-shot & 0 & 16.17 & 54.20 & 41.59 & 20.30 & 20.33 & 45.62 & 33.04 & -- & -- \\
\rowcolor{proudtableshade}
LoRA-FT~\cite{hu2022lora} & 61.0M & 58.03 & 77.63 & 44.39 & 67.40 & 61.77 & \textbf{58.22} & 61.24 & 76.55 & -18.78 \\
O-LoRA~\cite{wang2023olora} & 61.0M & 77.50 & 78.07 & 44.50 & 63.13 & 64.73 & \underline{58.16} & 64.35 & 78.02 & -13.99 \\
\rowcolor{proudtableshade}
MoELoRA~\cite{chen2024coin} & 72.8M & 70.07 & 77.70 & 44.69 & 50.03 & 54.03 & 57.34 & 58.98 & 75.08 & -14.63 \\
ModalPrompt~\cite{zeng2025modalprompt} & 41.0M & 51.07 & 87.27 & \underline{48.11} & 39.23 & 46.57 & 42.93 & 52.53 & 57.67 & \textbf{-0.15} \\
\rowcolor{proudtableshade}
CL-MoE~\cite{huai2025cl} & 267.7M & 66.33 & 77.00 & 44.78 & 51.87 & 53.53 & 57.42 & 58.49 & 74.19 & -15.56 \\
HiDe-LLaVA~\cite{guo2025hide} & 61.0M & \underline{84.03} & \underline{90.73} & 44.43 & 58.93 & 41.37 & 54.25 & 62.29 & 77.32 & -9.20 \\
\rowcolor{proudtableshade}
SEFE~\cite{chen2025sefe} & 61.0M & 80.83 & 78.00 & 47.01 & \textbf{69.63} & \underline{65.83} & 57.92 & \underline{66.54} & \underline{78.76} & -11.33 \\
\midrule
\rowcolor{proudtableresult}
\textbf{PROUD (Ours)} & \textbf{0.52M} & \textbf{89.70} & \textbf{92.93} & \textbf{57.68} & \underline{68.60} & \textbf{68.03} & 58.00 & \textbf{72.49} & \textbf{81.89} & \underline{-0.93} \\
\bottomrule
\end{tabularx}
\end{table*}

\begin{table*}[t]
\centering
\proudmaintabsetup
\prouddensetable
\setlength{\tabcolsep}{3pt}
\caption{Main results on the CoIN benchmark with LLaVA-v1.5-7B as the backbone. The best and second-best results are highlighted in \textbf{bold} and \underline{underline}, respectively.}
\label{tab:coin_benchmark}
\begin{tabularx}{\linewidth}{lcC{1.18}C{1.05}C{1.09}C{0.65}C{0.85}C{1.20}C{0.80}C{1.30}C{0.88}}
\toprule
\rowcolor{proudtableheadbg}
\textbf{Method} & \textbf{Params} & \textbf{ScienceQA} & \textbf{TextVQA} & \textbf{ImageNet} & \textbf{GQA} & \textbf{VizWiz} & \textbf{Grounding} & \textbf{VQAv2} & \textbf{OCR-VQA} & \proudtablehead{Last $\uparrow$} \\
\midrule
Finetune & 6.76B & 26.00 & 25.38 & 28.51 & 33.07 & 26.52 & 0.10 & 40.00 & 52.92 & 29.06 \\
\midrule
\rowcolor{proudtableshade}
CODA-Prompt~\cite{smith2023coda} & 165.0M & 58.15 & 50.16 & 24.04 & 54.33 & 48.94 & 17.83 & 55.86 & 54.42 & 45.46 \\
DualPrompt~\cite{wang2022dualprompt} & 21.3M & 56.40 & 47.12 & 34.96 & 42.03 & 44.14 & 12.01 & 54.43 & 53.36 & 43.05 \\
\rowcolor{proudtableshade}
L2P~\cite{wang2022learning} & 21.4M & 54.42 & 46.04 & 30.36 & 57.09 & 42.19 & 9.38 & 50.45 & 54.03 & 42.99 \\
MoELoRA~\cite{chen2024coin} & 349.9M & 47.34 & 32.91 & 38.73 & 37.15 & 42.48 & 0.97 & 42.77 & 57.50 & 37.48 \\
\rowcolor{proudtableshade}
Continual LLaVA~\cite{cao2024continual} & 0.75M & 58.67 & 49.99 & 57.66 & \textbf{62.53} & 42.32 & 16.25 & 64.33 & \textbf{74.91} & 53.33 \\
ModalPrompt~\cite{zeng2025modalprompt} & 41.0M & 68.42 & \underline{56.40} & 41.13 & \underline{61.11} & 50.13 & \underline{36.69} & \textbf{66.90} & 59.68 & 55.06 \\
\rowcolor{proudtableshade}
SEFE~\cite{chen2025sefe} & 340.8M & \underline{75.35} & \textbf{58.66} & \underline{83.10} & 54.25 & 48.85 & 16.75 & \underline{65.35} & \underline{66.25} & 58.57 \\
ProgLoRA~\cite{yu2025progressive} & 63.0M & 74.84 & 51.83 & \textbf{83.90} & 49.93 & \textbf{53.87} & 31.19 & 62.71 & 64.44 & \underline{59.09} \\
\midrule
\rowcolor{proudtableresult}
\textbf{PROUD (Ours)} & \textbf{0.52M} & \textbf{75.67} & 56.18 & 66.46 & 57.36 & \underline{52.56} & \textbf{51.92} & 62.68 & 61.41 & \textbf{60.53} \\
\bottomrule
\end{tabularx}
\end{table*}

\subsection{Isometric in-stage regularization}
\label{sec:isometric_training}

While Proposition~\ref{prop:optimal_subspace} identifies update directions with high initial benefit-to-cost ratios, unconstrained multi-step optimization along $P_t$ can still accumulate large displacements on historical representations. During stage $t$, we therefore freeze $W_{t-1}$ and the metric-orthonormal basis $P_t$, optimizing only $B_t \in \mathbb{R}^{m\times r}$ from $B_t^{(0)} = \mathbf{0}$ via the regularized objective:
\begin{equation}
    \min_{B_t}\; \mathcal{L}_t(B_t) = \mathcal{L}_t^{\mathrm{CE}}(W_{t-1} + B_t P_t^\top; \mathcal{D}_t) + \frac{\lambda}{2}\|B_t\|_F^2,
    \label{eq:stage_objective}
\end{equation}
where $\lambda \ge 0$ controls the regularization strength. Since the fixed basis satisfies $P_t^\top M_t P_t = I_r$, Eq.~\eqref{eq:stage_objective} yields an exact isometric identity and multi-stage drift bound.

\begin{proposition}[Isometric Regularization and Multi-Stage Drift Bound]
\label{prop:isometry}
Under $\Delta W_t = B_t P_t^\top$ with $P_t^\top M_t P_t = I_r$, stage-wise drift regularization is exact, and RMS token output drift on any earlier task $s<t$ satisfies:
\begin{align}
    q_t(\Delta W_t) &= \|B_t\|_F^2, \label{eq:isometry} \\
    \big\|(W_t - W_s)\widehat{C}_s^{1/2}\big\|_F &\le \sum_{k=s+1}^{t} \sqrt{k-1}\,\|B_k\|_F,
    \label{eq:drift_bound}
\end{align}
where $\|(W_t-W_s)\widehat{C}_s^{1/2}\|_F = \mathbb{E}_{h\sim\widehat{\mathcal{H}}_s}[\|(W_t-W_s)h\|_2^2]^{1/2}$ is the RMS token drift under the empirical task-$s$ token distribution $\widehat{\mathcal{H}}_s$ (proof in Appendix~\ref{app:isometry_proof}).
\end{proposition}

\noindent\textbf{Discussion.} Keeping $P_t$ fixed throughout stage $t$ ensures that multi-step updates remain within the selected subspace, reducing the quadratic drift penalty to an exact $\mathcal{O}(mr)$ Frobenius norm while halving Adam~\cite{kingma2015adam} optimizer states. Moreover, Eq.~\eqref{eq:drift_bound} shows that regularizing $\|B_t\|_F^2$ directly controls cumulative token displacement across stages, while merging $W_t \leftarrow W_{t-1} + B_t P_t^\top$ after each stage allows the effective adaptation rank to grow across tasks.

\noindent\textbf{Training and inference summary.} Algorithm~\ref{alg:proud} in the Appendix summarizes the complete stage-wise procedure. At the onset of stage $t$, PROUD forms the damped metric $M_t$ from retained statistics $C_{<t}$, estimates the preview gradient $\bar{G}_t$ on $\mathcal{B}_t$, and extracts the metric-orthonormal basis $P_t$ via Eq.~\eqref{eq:optimal_basis}. During training, only the outgoing factor $B_t$ is optimized under Eq.~\eqref{eq:stage_objective}. At stage end, the update $B_t P_t^\top$ is merged into $W_t$ and $\widehat{C}_t$ is accumulated into $C_{<t+1}$. At inference, the merged projector operates identically to the base model without task IDs, routing, or extra parameters.

\section{Experiments}
\label{sec:experiments}

\subsection{Experimental setup}
\label{subsec:exp_setup}
\noindent\textbf{Datasets and baselines.}
We evaluate PROUD under the replay-free protocols of CoIN~\cite{chen2024coin} across eight tasks~\cite{deng2009imagenet,goyal2017making,gurari2018vizwiz,hudson2019gqa,kazemzadeh2014referitgame,lu2022learn,mishra2019ocr,singh2019towards} and UCIT~\cite{guo2025hide,guo2025mcitlib} across six tasks with reduced domain overlap~\cite{gurari2020captioning,hendrycks2021many,li2024multimodal,lindstrom2022clevr,lu2021iconqa,young2014image}. We compare against prompt-based methods (L2P~\cite{wang2022learning}, DualPrompt~\cite{wang2022dualprompt}, CODA-Prompt~\cite{smith2023coda}, ModalPrompt~\cite{zeng2025modalprompt}) and LoRA/MoE methods (LoRA-FT~\cite{hu2022lora}, O-LoRA~\cite{wang2023olora}, Continual LLaVA~\cite{cao2024continual}, MoELoRA~\cite{chen2024coin}, CL-MoE~\cite{huai2025cl}, ProgLoRA~\cite{yu2025progressive}, HiDe-LLaVA~\cite{guo2025hide}, SEFE~\cite{chen2025sefe}). Unlike methods requiring tens to hundreds of millions of parameters, PROUD updates a single shared layer with 0.52M parameters on LLaVA and no routing.

\noindent\textbf{Training and evaluation.}
We use LLaVA-v1.5-7B~\cite{liu2024improved} and Qwen2.5-VL-7B-Instruct~\cite{bai2025qwen25vl}, adapting only the terminal projector layer $W$ ($r = 128$, $\lambda = 0.01$) for one epoch per task while freezing upstream modules and the decoder. We report final average accuracy (Last), incremental average accuracy (Avg), and backward transfer (BWT)~\cite{chen2024coin,guo2025hide}, with full details and stage matrices in the Appendix.

\subsection{Benchmark comparison}
\label{subsec:main_results}

\begin{figure}[t]
  \centering
  \includegraphics[width=\linewidth]{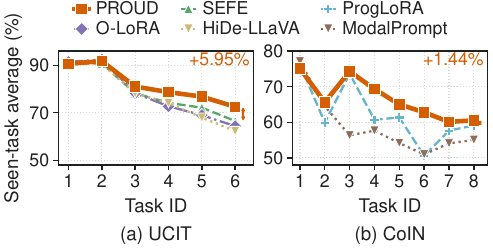}
  \caption{\textbf{Stage-wise continual performance on LLaVA-v1.5-7B.} Seen-task average accuracy across sequential learning stages on (a) UCIT and (b) CoIN. Annotations show the final-stage gain over the strongest displayed baseline.}
  \label{fig:continual_curves}
\end{figure}

As shown in Table~\ref{tab:ucit_benchmark}, PROUD achieves the highest final accuracy of 72.49\% on UCIT (\mbox{$72.53 \pm 0.26\%$} across three seeds), surpassing SEFE~\cite{chen2025sefe} (66.54\%) and O-LoRA~\cite{wang2023olora} (64.35\%) by 5.95 and 8.14 points. PROUD ranks first on four of six tasks---including gains of 9.57 points on VizWiz and 5.67 on ImageNet-R---and achieves the best overall Avg (81.89\%) with minimal forgetting ($-0.93\%$ BWT) using $117\times$ fewer parameters across stages in Figure~\ref{fig:continual_curves}(a).

On CoIN in Table~\ref{tab:coin_benchmark}, PROUD achieves \textbf{60.53\%} Last ($60.27 \pm 0.77\%$ across three seeds) under uniform $\lambda = 0.01$, surpassing ProgLoRA at 59.09\%, SEFE at 58.57\%, and ModalPrompt at 55.06\% with $79\times$ to $655\times$ fewer parameters. PROUD gains \textbf{15.23 points} on Grounding (51.92\% vs.\ 36.69\%), keeps ScienceQA stable (75.10\% to 75.67\%), and leads late-stage accuracy in Figure~\ref{fig:continual_curves}(b). Notably, PROUD retains 86.53\% on ImageNet through Stage~6; the high-exposure VQAv2 stage ($1.53\times$ mean exposure) may nudge shared object features toward coarse QA labels, and stronger Stage-7 regularization restores ImageNet to \textbf{77.49\%} and Last to \textbf{61.36\%} in Appendix~\ref{app:ablations}.

\begin{table*}[t]
\centering
\proudmaintabsetup
\caption{Continual instruction tuning results on the UCIT benchmark using Qwen2.5-VL-7B-Instruct as the backbone. Decoder baselines adapt LoRA modules across all FFN layers, whereas PROUD adapts only the terminal projector layer. The best and second-best results are highlighted in \textbf{bold} and \underline{underline}, respectively.}
\label{tab:qwen_ucit}
\prouddensetable
\setlength{\tabcolsep}{2.5pt}
\begin{tabularx}{\linewidth}{>{\raggedright\arraybackslash}p{0.21\linewidth}C{1.10}C{1.12}C{1.06}C{0.95}C{0.95}C{0.92}C{1.16}C{1.04}C{0.75}C{0.95}}
\toprule
\rowcolor{proudtableheadbg}
\textbf{Method} & \textbf{Params} & \textbf{ImgNet-R} & \textbf{ArxivQA} & \textbf{VizWiz} & \textbf{IconQA} & \textbf{CLEVR} & \textbf{Flickr30k} & \proudtablehead{\textbf{Last} $\uparrow$} & \proudtablehead{\textbf{Avg} $\uparrow$} & \proudtablehead{\textbf{BWT} $\uparrow$} \\
\midrule
LoRA-FT (Decoder) & 15.1M & 66.67 & 95.60 & 51.75 & \textbf{98.17} & 99.17 & \underline{60.96} & 78.72 & 85.65 & -8.59 \\
\rowcolor{proudtableshade}
MoELoRA (Decoder) & 95.2M & 65.97 & 95.53 & 51.19 & \underline{98.13} & \underline{99.50} & \textbf{61.05} & 78.56 & 85.67 & -8.80 \\
O-LoRA (Decoder) & 15.1M & \textbf{90.03} & \textbf{95.90} & \underline{54.14} & 97.10 & 99.40 & 60.58 & \underline{82.86} & \textbf{88.37} & \underline{-3.31} \\
\midrule
\rowcolor{proudtableresult}
\textbf{PROUD (Ours)} & \textbf{0.46M} & \underline{89.37} & \underline{95.73} & \textbf{65.87} & 96.07 & \textbf{99.60} & 58.35 & \textbf{84.17} & \underline{87.78} & \textbf{-0.06} \\
\bottomrule
\end{tabularx}
\end{table*}

\begin{figure*}[t]
  \centering
  \includegraphics[width=\linewidth]{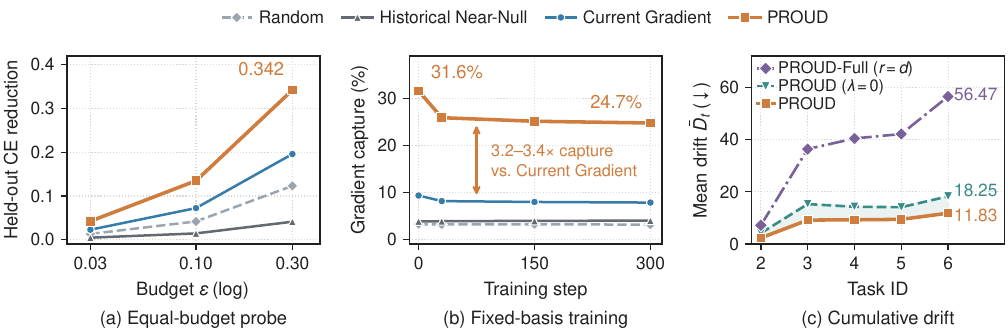}
  \caption{\textbf{Subspace direction quality and cumulative drift on UCIT.} (a) Held-out loss reduction under matched drift budgets and (b) whitened gradient capture during fixed-basis training on VizWiz-caption. (c) Mean interface displacement $\bar{D}_t$ on earlier tasks. PROUD achieves the largest loss reduction, sustains high gradient capture, and minimizes cumulative drift.}
  \label{fig:mechanisms}
\end{figure*}

To assess architectural generality, we evaluate PROUD on Qwen2.5-VL-7B in Table~\ref{tab:qwen_ucit}. PROUD achieves \textbf{84.17\%} Last ($84.26 \pm 0.23\%$ across three seeds) and near-zero forgetting at $-0.06\%$ BWT, outperforming O-LoRA at 82.86\%, LoRA-FT at 78.72\%, and MoELoRA at 78.56\% with $33\times$ to $207\times$ fewer per-task parameters. While O-LoRA accumulates 90.8M parameters for slightly higher Avg (88.37\% vs.\ 87.78\%), it suffers $-3.31\%$ BWT and loses 15.49 points on VizWiz after Stage~3 versus 0.02 for PROUD in Appendix~\ref{app:stage_matrices}.

\subsection{Controlled subspace and ablation studies}
\label{subsec:ablations}

\begin{table}[t]
\centering
\proudanalysistabsetup
\caption{Controlled subspace comparisons at the terminal projector layer on UCIT. Near-Null and Current Gradient isolate selection criteria from the cited methods. Best group results are in \textbf{bold}.}
\label{tab:main_cl}
\begin{tabularx}{\linewidth}{@{}>{\raggedright\arraybackslash}p{0.42\linewidth}C{1.05}C{0.95}C{0.95}C{1.05}@{}}
\toprule
\textbf{Method / Subspace} & \textbf{Params} & \textbf{Last} $\uparrow$ & \textbf{Avg} $\uparrow$ & \textbf{BWT} $\uparrow$ \\
\midrule
\multicolumn{5}{@{}l}{\textit{Unconstrained and full-space adaptation}} \\
Seq.\ Full-Rank FT & 16.78M & 57.48 & 67.84 & -24.80 \\
Seq.\ LoRA & 1.05M & 47.63 & 56.01 & -25.17 \\
Full-Space Soft Reg & 16.78M & 71.27 & 80.49 & -7.60 \\
PROUD-Full ($r=d$) & 16.78M & \textbf{72.28} & \textbf{82.40} & \textbf{-0.84} \\
\midrule
\multicolumn{5}{@{}l}{\textit{Rank-matched fixed subspaces ($r=128$)}} \\
Random Basis & 0.52M & 65.40 & 77.31 & -7.31 \\
Historical Near-Null~\cite{liang2024inflora,saha2021gpm} & 0.52M & 61.01 & 74.78 & -1.67 \\
Current Gradient~\cite{wang2024loraga} & 0.52M & 67.89 & 80.09 & -3.96 \\
\rowcolor{proudtableresult}
\textbf{PROUD (Ours)} & \textbf{0.52M} & \textbf{72.49} & \textbf{81.89} & \textbf{-0.93} \\
\bottomrule
\end{tabularx}
\end{table}

\begin{table}[t]
\centering
\proudanalysistabsetup
\renewcommand{\arraystretch}{1.15}
\caption{Historical-metric ablations at the terminal projector layer on UCIT. Spec./Orient. retain the eigenvalues/eigenvectors of the included history; Accum. pools stages.}
\label{tab:ablations}
\begin{tabularx}{\linewidth}{@{}>{\raggedright\arraybackslash}p{0.22\linewidth}C{0.94}C{1.05}C{1.12}C{0.95}C{0.91}C{1.03}@{}}
\toprule
\textbf{Variant} & \textbf{Spec.} & \textbf{Orient.} & \textbf{Accum.} & \textbf{Last} $\uparrow$ & \textbf{Avg} $\uparrow$ & \textbf{BWT} $\uparrow$ \\
\midrule
No history & \proudxmark & \proudxmark & \proudxmark & 53.39 & 68.52 & -28.13 \\
Rotated & \proudcmark & \proudxmark & \proudcmark & 57.64 & 73.56 & -23.54 \\
Prev.\ stage & \proudcmark & \proudcmark & \proudxmark & 68.40 & 81.05 & -7.64 \\
\rowcolor{proudtableresult}
\textbf{PROUD} & \proudcmark & \proudcmark & \proudcmark & \textbf{72.49} & \textbf{81.89} & \textbf{-0.93} \\
\bottomrule
\end{tabularx}
\end{table}

\edef\proudsavedtopnumber{\the\value{topnumber}}
\setcounter{topnumber}{1}
\begin{figure}[t]
  \centering
  \includegraphics[width=\linewidth]{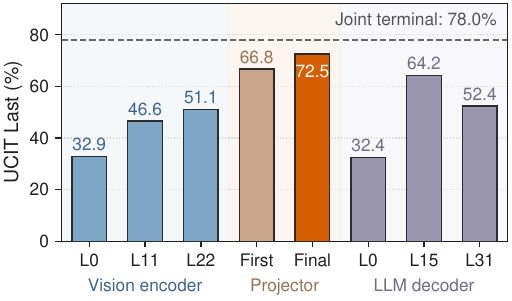}
  \caption{\textbf{Impact of adaptation placement on UCIT.} Final accuracy across vision-encoder, projector, and language-decoder sites under a matched trainable parameter budget. Terminal-projector adaptation approaches its joint-training reference shown by the dashed line and outperforms encoder or decoder tuning.}
  \label{fig:locus}
  \par\medskip
\begin{minipage}{\linewidth}
\centering
\proudanalysistabsetup
\captionof{table}{Efficiency and routed-expert comparisons on UCIT. Memory is in MiB, training time in hours, and speed in tokens/s. Latency (ms/sample) measures first-token generation in (a) and additional routing and dispatch in (b).}
\label{tab:resources}
\begin{tabularx}{\linewidth}{@{}>{\raggedright\arraybackslash}p{0.285\linewidth}C{1.04}C{1.07}C{0.99}C{0.95}C{0.95}@{}}
\toprule
\multicolumn{6}{@{}l}{\textbf{(a) Training and merged inference ($r=128$)}} \\
\cmidrule(lr){1-6}
\textbf{Method} & \textbf{Adapter} & \textbf{Adam} & \textbf{Time} & \textbf{Latency} & \textbf{Speed} \\
\midrule
Base MLLM & 0 & 0 & -- & 88.85 & 43.80 \\
\midrule
Full-Space Soft Reg & 64.0 & 128.0 & 5.16 & 89.23 & 44.07 \\
Seq.\ LoRA & \textbf{4.0} & 8.0 & \textbf{4.85} & 89.10 & 44.07 \\
\rowcolor{proudtableresult}
\textbf{PROUD (Ours)} & \textbf{4.0} & \textbf{4.0} & 4.90 & 89.08 & 44.49 \\
\midrule
\end{tabularx}\par\nointerlineskip
\setlength{\tabcolsep}{1pt}
\begin{tabularx}{\linewidth}{@{}>{\raggedright\arraybackslash}p{0.255\linewidth}C{0.95}C{0.98}C{1.67}C{0.7}C{0.7}@{}}
\multicolumn{6}{@{}l}{\textbf{(b) Shared layer and routed experts ($r=32$)}} \\
\cmidrule(lr){1-6}
\textbf{Method} & \textbf{Routing} & \textbf{Modules} & \textbf{Overhead (ms)} & \textbf{Last} $\uparrow$ & \textbf{Avg} $\uparrow$ \\
\midrule
Per-task LoRA & CLIP & 6 & +34.93 & 62.91 & 64.95 \\
Per-task LoRA & Oracle & 6 & +0.02 & 62.94 & 64.96 \\
\rowcolor{proudtableresult}
\textbf{PROUD (Ours)} & None & 1 & +0.00 & \textbf{68.87} & \textbf{78.27} \\
\bottomrule
\end{tabularx}
\end{minipage}

\end{figure}

\begin{figure}[t]
  \centering
  \includegraphics[width=\linewidth]{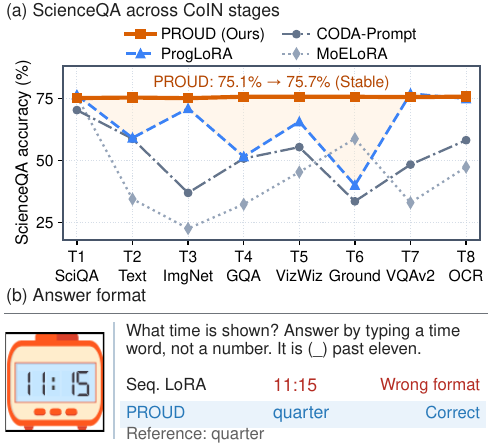}
  \caption{\textbf{Answer-format retention on CoIN and IconQA.} (a) ScienceQA accuracy across eight CoIN stages. PROUD avoids the severe accuracy fluctuations of baselines. (b) Qualitative example where unconstrained adaptation violates the requested word format (\texttt{11:15}), whereas PROUD preserves \texttt{quarter}.}
  \label{fig:qualitative_main}
\end{figure}

\noindent\textbf{Controlled subspace comparison.}
To isolate the impact of subspace selection from architectural differences, Table~\ref{tab:main_cl} compares PROUD against unconstrained adaptation, two full-space variants, and three rank-matched subspaces at the shared terminal projector layer. All fixed-basis baselines are normalized under the same historical metric $M_t$ and trained via Eq.~\eqref{eq:stage_objective}. Under this controlled setup, isolating historical null-space projection (Historical Near-Null~\cite{liang2024inflora,saha2021gpm}) limits forgetting but severely restricts plasticity, trailing even a Random Basis. Conversely, greedy gradient alignment (Current Gradient~\cite{wang2024loraga}) captures task gradients but collides with dominant historical directions ($-3.96\%$ BWT). By jointly optimizing the benefit-to-cost ratio, PROUD outperforms both single-criterion subspaces across all metrics (72.49\% Last, $-0.93\%$ BWT) and even surpasses Full-Space Soft Reg. Compared with PROUD-Full ($r = d$), PROUD updates $32\times$ fewer parameters, cuts adapter and optimizer memory by $24\times$, and reduces cumulative drift by $4.8\times$ in Figure~\ref{fig:mechanisms}(c), with Eq.~\eqref{eq:stage_objective} cutting drift by $35.2\%$ over $\lambda = 0$. Truncation filters low-benefit directions, raising Last from 72.28\% to 72.49\% on LLaVA and 82.85\% to 84.17\% on Qwen2.5-VL, with robustness across task orderings, ranks, preview budgets, and instruction paraphrases in Appendix~\ref{app:ablations}.

\noindent\textbf{Local probing and trajectories.}
To test Proposition~\ref{prop:optimal_subspace} at Stage~$2\to3$, we take an analytical step along each basis under matched drift $q_3(\Delta W)=\varepsilon^2$ and evaluate loss drop on a held-out split (Appendix~\ref{app:ablations}). Historical Near-Null captures less than $3.5\%$ of whitened gradient energy, whereas PROUD captures $39.77\%$ on VizWiz and $86.48\%$ on IconQA, yielding the largest loss drop across budgets in Figure~\ref{fig:mechanisms}(a). Across 300 steps in Figure~\ref{fig:mechanisms}(b), Current Gradient suffers severe step-size attenuation once normalized under $M_t$ because its directions overlap with high-energy historical features, whereas PROUD sustains over $3\times$ higher whitened gradient capture under stationary inputs.

\noindent\textbf{Metric and placement ablations.}
To verify our historical metric and terminal-layer placement, we ablate metric components in Table~\ref{tab:ablations} and adaptation sites in Figure~\ref{fig:locus}. Removing historical statistics ($C_{<t} = \mathbf{0}$), randomly rotating the eigenvectors of $C_{<t}$ while preserving its spectrum, or using only previous-stage statistics all degrade retention, confirming that PROUD relies on full historical token directions. Weighting projector outputs by downstream loss-gradient sensitivity in Appendix~\ref{app:ablations} yields comparable accuracy ($72.38\%$ vs.\ $72.49\%$) at higher cost, supporting the input-only metric. Across adaptation sites under a matched 0.52M parameter budget in Figure~\ref{fig:locus} and Appendix~\ref{app:ablations}, adapting vision-encoder or language-decoder V-projections consistently trails the terminal projector layer. Although the first projector layer also receives stationary inputs, it trails the terminal layer by 5.69 points because the intermediate GELU and second projection break the exact correspondence between its local quadratic metric and final visual-token drift.

\subsection{Further analysis}
\label{subsec:further_analysis}
\noindent\textbf{Format-retention and qualitative analysis.} To examine how curbing interface drift preserves task conventions beyond aggregate accuracy, we analyze answer-format stability in Figure~\ref{fig:qualitative_main}. Updating multi-layer decoders on lowercase VQA tasks often overwrites uppercase multiple-choice conventions on ScienceQA, causing severe accuracy fluctuations in Figure~\ref{fig:qualitative_main}(a), whereas PROUD freezes the decoder and keeps ScienceQA stable from $75.10\%$ to $75.67\%$ across all eight stages. Similarly, on IconQA in Figure~\ref{fig:qualitative_main}(b), unconstrained adaptation replaces the requested time word with digits (\texttt{11:15}), whereas PROUD preserves strict instruction-format adherence. Across all 48 IconQA test questions requesting a time word rather than digits, Seq.\ LoRA violates this format on $89.6\%$ of samples (43/48), whereas PROUD reduces the violation rate to $0.0\%$ (0/48), with the full format audit and qualitative examples provided in Appendix~\ref{app:stage_matrices}.

\noindent\textbf{Training and inference efficiency.} Table~\ref{tab:resources}(a) profiles training cost and merged-inference efficiency. Freezing $P_t$ halves Adam-state memory from 8.0 to 4.0\,MiB relative to Seq.\ LoRA while requiring only a fixed 64\,MiB FP32 training buffer for $C_{<t}$ that is discarded at inference, and stage-onset SVD plus statistics collection take only $1.5\%$ of training time (Appendix~\ref{app:efficiency}). After weight merging, PROUD matches the base model in inference latency and generation speed with zero added parameters.

\noindent\textbf{Comparison with routed LoRA experts.} To test whether a shared interface can outperform expert expansion, \mbox{Table~\ref{tab:resources}(b)} compares PROUD with routed per-task LoRA experts at matched rank ($r = 32$). PROUD reaches 68.87\% Last and 78.27\% Avg, surpassing both CLIP-routed~\cite{radford2021learning} (62.91\% Last, 64.95\% Avg) and oracle-routed (62.94\% Last, 64.96\% Avg) experts by nearly 6 points in Last and over 13 points in Avg. Even without routing errors, isolating tasks into separate experts limits knowledge sharing, whereas adapting a shared interface enables positive cross-task transfer.

\setcounter{topnumber}{\proudsavedtopnumber}

\section{Conclusion}
\label{sec:conclusion}

We studied replay-free Multimodal Continual Instruction Tuning under a fixed-capacity constraint, motivated by the observation that sequential updates at the shared vision--language projector induce severe interface drift. To balance plasticity and stability, we proposed PROUD, a continual subspace optimization framework at the coordinate-stationary final projector layer. By coupling retained input statistics with initial task gradients via metric-whitened SVD, PROUD analytically selects the optimal benefit-to-cost subspace at task onset and enforces exact isometric drift regularization during training, followed by lossless weight merging with zero inference overhead. Extensive experiments on UCIT and CoIN across multiple MLLM backbones validate its effectiveness and efficiency.

\FloatBarrier

{\small
\bibliographystyle{ieeenat_fullname}
\bibliography{references}
}

\onecolumn
\microtypesetup{activate=false}
\setcounter{equation}{0}
\setcounter{table}{0}
\setcounter{figure}{0}
\renewcommand{\theequation}{S\arabic{equation}}
\renewcommand{\thetable}{S\arabic{table}}
\renewcommand{\thefigure}{S\arabic{figure}}
\renewcommand{\theHequation}{supp.\arabic{equation}}
\renewcommand{\theHtable}{supp.\arabic{table}}
\renewcommand{\theHfigure}{supp.\arabic{figure}}

\begin{center}
  {\Large\bfseries Supplementary Material\par}
  \vspace{1em}
\end{center}
\appendix
\section*{Appendix}

In this appendix, we provide complete mathematical proofs, implementation verification protocols, dataset and baseline specifications, and additional experimental analyses for \textbf{PROUD}:
\begin{itemize}
    \item \textbf{Section~\ref{app:proofs}} provides the complete proofs of Proposition~\ref{prop:optimal_subspace}, the generalized eigenvalue equivalence, the isometric regularization identity in Eq.~\eqref{eq:isometry}, multi-stage displacement bounds, and the diagnostic decomposition (\textbf{Section~\ref{app:diagnostic_decomposition}}).
    \item \textbf{Section~\ref{app:datasets}} describes the benchmark protocols, default training settings, compared baselines, controlled subspace formulations (\textbf{Section~\ref{app:controlled_baselines}}), and the continual-stage procedure (\textbf{Section~\ref{app:algorithm}}).
    \item \textbf{Section~\ref{app:metrics}} defines the continual evaluation metrics ($\text{Last}$, $\text{Avg}$, $\text{BWT}$).
    \item \textbf{Section~\ref{app:stage_matrices}} reports the complete stage-by-stage result matrices for UCIT and CoIN.
    \item \textbf{Section~\ref{app:ablations}} reports extended ablations on rank and learning-rate grids, three-seed replication, test-instruction variation, output-sensitivity controls, adaptation placements, joint-training references, 300-step gradient capture trajectories, same-layer and parameter-matched controls, segmented stage-boundary stress tests, and the Qwen2.5-VL PROUD-Full stage matrix.
    \item \textbf{Section~\ref{app:efficiency}} provides detailed stage-onset matrix decomposition timings and training/inference resource accounting.
    \item \textbf{Section~\ref{app:numerical_checks}} details the numerical verification checklist for CPU FP64 decompositions and FP32 training and weight merging.
\end{itemize}

\section{Mathematical proofs and drift analysis}
\label{app:proofs}

\subsection{Proof of Proposition~\ref{prop:optimal_subspace} (optimal update subspace)}
\label{app:optimal_subspace_proof}

We restate and prove Proposition~\ref{prop:optimal_subspace} from Section~\ref{sec:subspace_selection}. Recall the constrained local subspace optimization problem at stage $t$ (assuming non-vanishing preview gradient $\bar{G}_t \neq \mathbf{0}$):
\begin{equation}
    \max_{\Delta W \in \mathbb{R}^{m\times d}} -\langle \bar{G}_t, \Delta W \rangle_F \quad \text{s.t.} \quad \operatorname{rank}(\Delta W) \le r, \;\; q_t(\Delta W) = \operatorname{tr}(\Delta W M_t \Delta W^\top) \le \varepsilon^2.
    \label{eq:app_prob}
\end{equation}

\begin{proof}
Since $C_{<t} \succeq 0$ and $\rho > 0$, the damped historical metric $M_t = C_{<t} + \rho I_d$ is symmetric positive-definite ($M_t \succ 0$) and admits a unique Cholesky factorization $M_t = L_t L_t^\top$, where $L_t \in \mathbb{R}^{d\times d}$ is lower triangular and nonsingular.

\noindent\textbf{Step 1: Metric whitening and coordinate transformation.}
Define the invertible linear change of variables:
\begin{equation}
    X = \Delta W L_t \in \mathbb{R}^{m\times d} \quad \Longleftrightarrow \quad \Delta W = X L_t^{-1}.
    \label{eq:app_change_var}
\end{equation}
Because $L_t$ is full rank, $\operatorname{rank}(X) = \operatorname{rank}(\Delta W) \le r$. Substituting $\Delta W = X L_t^{-1}$ into the displacement quadratic form yields an isotropic Frobenius norm constraint:
\begin{equation}
    q_t(\Delta W) = \operatorname{tr}\big(X L_t^{-1} (L_t L_t^\top) L_t^{-\top} X^\top\big) = \operatorname{tr}(X X^\top) = \|X\|_F^2 \le \varepsilon^2.
\end{equation}
Defining the metric-whitened preview gradient $K_t = \bar{G}_t L_t^{-\top} \in \mathbb{R}^{m\times d}$, the first-order objective becomes:
\begin{equation}
    -\langle \bar{G}_t, \Delta W \rangle_F = -\operatorname{tr}(\bar{G}_t L_t^{-\top} X^\top) = -\langle K_t, X \rangle_F.
\end{equation}
Thus, Problem~\eqref{eq:app_prob} is strictly equivalent to the Euclidean low-rank optimization problem:
\begin{equation}
    \max_{X \in \mathbb{R}^{m\times d}} -\langle K_t, X \rangle_F \quad \text{s.t.} \quad \operatorname{rank}(X) \le r, \;\; \|X\|_F \le \varepsilon.
    \label{eq:app_euclid}
\end{equation}

\noindent\textbf{Step 2: Optimal rank-$r$ subspace via Ky Fan's maximum principle.}
Any matrix $X \in \mathbb{R}^{m\times d}$ with $\operatorname{rank}(X) \le r$ can be factored as $X = B V^\top$, where $V \in \mathbb{R}^{d\times r}$ has orthonormal columns ($V^\top V = I_r$) and $B \in \mathbb{R}^{m\times r}$, with $\|X\|_F^2 = \operatorname{tr}(B V^\top V B^\top) = \|B\|_F^2 \le \varepsilon^2$. For any fixed orthonormal basis $V$ with $K_t V \neq \mathbf{0}$, maximizing $-\langle K_t, B V^\top \rangle_F = -\langle K_t V, B \rangle_F$ over $\|B\|_F \le \varepsilon$ is solved by the Cauchy--Schwarz inequality:
\begin{equation}
    \max_{\|B\|_F \le \varepsilon} -\langle K_t V, B \rangle_F = \varepsilon \|K_t V\|_F, \qquad \text{attained at} \quad B^\star(V) = -\varepsilon \frac{K_t V}{\|K_t V\|_F}.
\end{equation}
Consequently, finding the optimal rank-$r$ subspace reduces to maximizing $\|K_t V\|_F^2 = \operatorname{tr}(V^\top K_t^\top K_t V)$ over $V^\top V = I_r$. Let $K_t = U_t \Sigma_t V_t^\top$ be the SVD of $K_t$ with singular values $\sigma_1(K_t) \ge \dots \ge \sigma_{\min(m,d)}(K_t) \ge 0$. By Ky Fan's maximum principle, $\operatorname{tr}(V^\top K_t^\top K_t V)$ is maximized when $V = V_{t,r} \in \mathbb{R}^{d\times r}$ consists of the top-$r$ right singular vectors of $K_t$, achieving $\max_{V^\top V = I_r} \|K_t V\|_F^2 = \sum_{i=1}^r \sigma_i^2(K_t)$.

\noindent\textbf{Step 3: Inverse mapping to original input coordinates.}
Mapping $X^\star = B^\star V_{t,r}^\top$ back to the original parameter space via Eq.~\eqref{eq:app_change_var} gives:
\begin{equation}
    \Delta W^\star = X^\star L_t^{-1} = B^\star \big(L_t^{-\top} V_{t,r}\big)^\top = B^\star P_t^\top, \qquad \text{where} \quad P_t \triangleq L_t^{-\top} V_{t,r} \in \mathbb{R}^{d\times r}.
\end{equation}
Verifying metric orthonormality directly:
\begin{equation}
    P_t^\top M_t P_t = V_{t,r}^\top L_t^{-1} (L_t L_t^\top) L_t^{-\top} V_{t,r} = V_{t,r}^\top V_{t,r} = I_r.
\end{equation}
Finally, since $K_t V_{t,r} = \bar{G}_t L_t^{-\top} V_{t,r} = \bar{G}_t P_t$, the criterion $\max_{V^\top V = I_r} \|K_t V\|_F^2$ translates into $\max_{P^\top M_t P = I_r} \|\bar{G}_t P\|_F^2 = \sum_{i=1}^r \sigma_i^2(K_t)$, and $B^\star(V_{t,r})$ yields the analytical equal-budget probe step $B_{\mathrm{probe}} = -\varepsilon \bar{G}_t P_t / \|\bar{G}_t P_t\|_F$.
\end{proof}

\subsection{Equivalence to the generalized eigenvalue problem and isometric identity}
\label{app:isometry_proof}

\noindent\textbf{Generalized eigenvalue formulation.}
Forming the Lagrangian for a single direction $p_i \in \mathbb{R}^d$ in $\max_{p_i^\top M_t p_i = 1} \|\bar{G}_t p_i\|_2^2$ gives $\mathcal{J}(p_i, \nu_i) = p_i^\top \bar{G}_t^\top \bar{G}_t p_i - \nu_i(p_i^\top M_t p_i - 1)$. Setting $\nabla_{p_i}\mathcal{J} = \mathbf{0}$ yields:
\begin{equation}
    \bar{G}_t^\top \bar{G}_t p_i = \nu_i M_t p_i \quad \Longleftrightarrow \quad \big(L_t^{-1} \bar{G}_t^\top \bar{G}_t L_t^{-\top}\big) v_i = \nu_i v_i,
\end{equation}
where $v_i = L_t^\top p_i$ is the $i$-th right singular vector of $K_t = \bar{G}_t L_t^{-\top}$, and $\nu_i = \sigma_i^2(K_t) = \|\bar{G}_t p_i\|_2^2 / (p_i^\top M_t p_i)$ is the directional benefit-to-cost ratio.

\noindent\textbf{Isometric regularization identity.}
Under the parameterization $\Delta W = W - W_{t-1} = B_t P_t^\top$ with $P_t^\top M_t P_t = I_r$, substituting $\Delta W$ into $q_t(\cdot)$ and applying the cyclic property of the trace yields Eq.~\eqref{eq:isometry}:
\begin{equation}
    q_t(W - W_{t-1}) = \operatorname{tr}\Big(B_t \underset{= I_r}{(P_t^\top M_t P_t)} B_t^\top\Big) = \operatorname{tr}(B_t B_t^\top) = \|B_t\|_F^2.
\end{equation}

\noindent\textbf{Multi-stage drift relation.}
While $\|B_k\|_F^2 = \operatorname{tr}(\Delta W_k M_k \Delta W_k^\top)$ controls single-stage displacement at stage $k$, how does adaptation across subsequent stages $k \in \{s+1, \dots, t\}$ affect a specific historical task $s$? Since $C_{<k} = \frac{1}{k-1}\sum_{j=1}^{k-1}\widehat{C}_j$ with $\widehat{C}_j \succeq 0$, we have $M_k \succeq C_{<k} \succeq \frac{1}{k-1}\widehat{C}_s$ for any $s < k$. Thus, the single-stage drift on task $s$ satisfies $\mathbb{E}_{h\sim \widehat{\mathcal{H}}_s}[\|\Delta W_k h\|_2^2] = \operatorname{tr}(\Delta W_k \widehat{C}_s \Delta W_k^\top) \le (k-1)\|B_k\|_F^2$, and over $t-s$ stages the net displacement obeys the triangle inequality $\|(W_t - W_s)\widehat{C}_s^{1/2}\|_F \le \sum_{k=s+1}^t \sqrt{k-1}\,\|B_k\|_F$.

\noindent\textbf{Conditional historical loss bound.}
Because the downstream decoder and projector bias are frozen, each historical example's teacher-forced cross-entropy $\ell_{s,i}$ is a fixed function of its projected visual-token sequence $z_{i,j}=Wh_{i,j}+b$. Let $\widehat{\mathcal{L}}_s^{\mathrm{CE}}=n_s^{-1}\sum_i\ell_{s,i}$ denote loss on the same empirical image set used in Eq.~\eqref{eq:stage_cov}. Assume that, for the compared weights, these loss maps are uniformly $L_s$-Lipschitz with respect to the sampled token-RMS distance $d_i(W,W')=\big[|\mathcal{J}_i|^{-1}\sum_{j\in\mathcal{J}_i}\|(W-W')h_{i,j}\|_2^2\big]^{1/2}$. This condition links per-example cross-entropy sensitivity to the empirical token-RMS metric induced by $\widehat{C}_s$. The triangle inequality, Jensen's inequality, and Eq.~\eqref{eq:drift_bound} then give
\begin{equation*}
\begin{aligned}
\big|\widehat{\mathcal{L}}_s^{\mathrm{CE}}(W_t)-\widehat{\mathcal{L}}_s^{\mathrm{CE}}(W_s)\big|
&\le \frac{L_s}{n_s}\sum_{i=1}^{n_s} d_i(W_t,W_s) \\
&\le L_s\left[\frac{1}{n_s}\sum_{i=1}^{n_s}d_i(W_t,W_s)^2\right]^{1/2}
=L_s\big\|(W_t-W_s)\widehat{C}_s^{1/2}\big\|_F \\
&\le L_s\sum_{k=s+1}^{t}\sqrt{k-1}\,\|B_k\|_F.
\end{aligned}
\end{equation*}

\subsection{Diagnostic decomposition of projector adaptation}
\label{app:diagnostic_decomposition}

Let $h = \phi(I) \in \mathbb{R}^d$ denote an upstream visual-token feature from image $I$ to the adapted layer. Substituting $\Delta W_t = B_t P_t^\top$ into the first-order task loss reduction and expected historical output perturbation $\|\Delta W_t h\|_2^2$ exposes the coupled trade-off in Eq.~\eqref{eq:tradeoff}. The upstream features $\phi(I)$, subspace basis $P_t$, and outgoing factor $B_t$ give rise to three distinct obstacles:
\begin{enumerate}
    \item \textbf{Feature non-stationarity in $\phi(I)$.} Updating multi-layer backbones or upstream adapters continually shifts the mapping $\phi(I)$ across stages, invalidating accumulated second-moment statistics $\mathbb{E}[\phi(I)\phi(I)^\top]$ without data replay.
    \item \textbf{Onset directional mismatch in $P_t$.} Selecting $P_t$ solely from the minor eigenvectors of $\mathbb{E}[\phi(I)\phi(I)^\top]$ suppresses the historical-drift term in Eq.~\eqref{eq:tradeoff} but can yield vanishing projected gradients $\bar{G}_t P_t \approx \mathbf{0}$ in the plasticity term. Selecting $P_t$ solely from $\bar{G}_t$ maximizes initial descent but can inflate historical interference $P_t^\top \mathbb{E}[\phi(I)\phi(I)^\top] P_t$.
    \item \textbf{In-stage magnitude drift in $B_t$.} Even when $P_t$ is well-chosen at task onset, unconstrained multi-step optimization of $B_t$ amplifies $\|B_t\|_F$, causing cumulative displacement in Eq.~\eqref{eq:tradeoff} to grow across training steps.
\end{enumerate}
PROUD addresses these components through a coordinate-stationary terminal projector interface (Section~\ref{sec:stationary_metric}), budgeted onset selection of $P_t$ (Section~\ref{sec:subspace_selection}), and isometric regularization of $B_t$ (Section~\ref{sec:isometric_training}). The full trace expansion and cumulative bound are given in Section~\ref{app:isometry_proof}.

\section{Benchmark protocols and compared baselines}
\label{app:datasets}

\subsection{Datasets and training exposure budgets}
\begin{itemize}
    \item \textbf{UCIT}~\citep{guo2025hide} evaluates six sequential tasks in the order: \emph{ImageNet-R}~\citep{hendrycks2021many} $\to$ \emph{ArxivQA}~\citep{li2024multimodal} $\to$ \emph{VizWiz-caption}~\citep{gurari2020captioning} $\to$ \emph{IconQA}~\citep{lu2021iconqa} $\to$ \emph{CLEVR-Math}~\citep{lindstrom2022clevr} $\to$ \emph{Flickr30k}~\citep{young2014image}. The one-epoch training budget comprises $24{,}000$, $39{,}984$, $39{,}984$, $29{,}856$, $39{,}984$, and $39{,}984$ training exposure instances per task ($213{,}792$ total sample exposures).
    \item \textbf{CoIN}~\citep{chen2024coin} evaluates eight sequential tasks in the order: \emph{ScienceQA}~\citep{lu2022learn} $\to$ \emph{TextVQA}~\citep{singh2019towards} $\to$ \emph{ImageNet}~\citep{deng2009imagenet} $\to$ \emph{GQA}~\citep{hudson2019gqa} $\to$ \emph{VizWiz}~\citep{gurari2018vizwiz} $\to$ \emph{RefCOCO Grounding}~\citep{kazemzadeh2014referitgame} $\to$ \emph{VQAv2}~\citep{goyal2017making} $\to$ \emph{OCR-VQA}~\citep{mishra2019ocr}, with a total one-epoch budget of $573{,}600$ sample exposures across the eight tasks. The VQAv2 training set contains $82{,}783$ examples and yields $82{,}752$ recorded exposures, $1.53\times$ the mean of the first six stages ($54{,}256$); its $214{,}354$ examples belong to the test split.
\end{itemize}

\noindent\textbf{Backbones and default training settings.}
We adopt LLaVA-v1.5-7B~\citep{liu2024improved} as the primary backbone and Qwen2.5-VL-7B-Instruct~\citep{bai2025qwen25vl} to test architectural generality. PROUD adapts only the final linear layer $W$ of the projector ($4096 \times 4096$ in LLaVA; $3584 \times 5120$ in Qwen2.5-VL), with the upstream modules and language decoder frozen. By default, we set $r=128$, learning rate $8\times10^{-5}$, $\lambda=10^{-2}$, and $\rho=\gamma s_0$ with $\gamma=10^{-2}$. Here $s_0=\operatorname{tr}(\widehat{C}_{\mathcal{B}_1})/d$, where $\widehat{C}_{\mathcal{B}_1}$ is the token second moment on the Stage-1 preview set before training. At each stage onset, $\bar{G}_t$ is estimated on $|\mathcal{B}_t|=256$ examples; at stage end, $\widehat{C}_t$ is estimated on at most $n_s=512$ examples with 32 tokens per image. Each task is trained for one epoch using Adam~\citep{kingma2015adam} with batch size 16. Projector training, second-moment storage, and weight merging use FP32. Cholesky factorization, SVD, and triangular solves run on CPU in FP64 for numerical accuracy; the resulting basis $P_t$ is then cast to FP32 for training and merging. Continual evaluation metrics are defined in Section~\ref{app:metrics}.

PROUD results in Tables~\ref{tab:ucit_benchmark}--\ref{tab:qwen_ucit} use seed 42. Table~\ref{tab:app_multiseed} reports three-seed results for the default UCIT and CoIN configurations.

\subsection{Compared baselines and controlled subspace formulations}
\label{app:controlled_baselines}

\noindent\textbf{Benchmark baselines.}
On CoIN and UCIT, we compare against: (1) \textbf{Finetune / LoRA-FT}, which sequentially updates the model without forgetting constraints; (2) \textbf{L2P}~\citep{wang2022learning}, \textbf{DualPrompt}~\citep{wang2022dualprompt}, \textbf{CODA-Prompt}~\citep{smith2023coda}, and \textbf{ModalPrompt}~\citep{zeng2025modalprompt}, which maintain task-invariant and task-specific prompt pools; (3) \textbf{MoELoRA}~\citep{chen2024coin}, \textbf{Continual LLaVA}~\citep{cao2024continual}, \textbf{CL-MoE}~\citep{huai2025cl}, \textbf{ProgLoRA}~\citep{yu2025progressive}, and \textbf{HiDe-LLaVA}~\citep{guo2025hide}, which expand or route task-specialized LoRA experts; and (4) \textbf{O-LoRA}~\citep{wang2023olora} and \textbf{SEFE}~\citep{chen2025sefe}, which enforce orthogonal adapter subspaces or weight-update regularization.

\noindent\textbf{Trainable-parameter accounting.}
In both benchmark tables, Params counts the scalar parameters updated during one task, including the visual projector when trainable and excluding frozen backbones, frozen historical modules, and non-trainable buffers. This is neither cumulative model storage nor the expert count activated by a single input. Counts follow each benchmark's adapter ranks and trainable modules. UCIT configurations follow MCITlib~\citep{guo2025mcitlib}. CoIN configurations follow the corresponding method implementations and publications.

\noindent\textbf{Qwen2.5-VL adaptation and parameter counts.}
In Table~\ref{tab:qwen_ucit}, decoder baselines adapt 84 LoRA modules across all 28 FFN layers. O-LoRA adds 15.14M trainable parameters per task, accumulating 90.83M parameters across the six UCIT tasks. PROUD updates only 0.46M parameters per task at the terminal projector layer and merges them after each task with zero parameter growth. In Qwen2.5-VL, this adaptation site corresponds to the terminal linear layer of the PatchMerger projector.

\noindent\textbf{Controlled interface baselines.}
To isolate directional selection at the exact same terminal projector layer $W \in \mathbb{R}^{m\times d}$, all fixed-basis baselines normalize their candidate basis $Q \in \mathbb{R}^{d\times r}$ via $P_Q = Q(Q^\top M_t Q)^{-1/2}$ ($P_Q^\top M_t P_Q = I_r$), parameterize $\Delta W = B_t P_Q^\top$, and minimize the exact same regularized loss in Eq.~\eqref{eq:stage_objective}:
\begin{enumerate}
    \item \textbf{Random Basis:} $Q \sim \mathcal{N}(0, I_{d\times r})$, testing whether low dimensionality plus metric regularization suffices.
    \item \textbf{Historical Near-Null:} $Q$ is formed by the bottom-$r$ eigenvectors of $C_{<t}$, isolating a historical null-space selection criterion inspired by GPM and InfLoRA~\citep{liang2024inflora,saha2021gpm}.
    \item \textbf{Current Gradient:} $Q$ is formed by the top-$r$ right singular vectors of the unwhitened preview gradient $\bar{G}_t$, isolating greedy gradient selection~\citep{wang2024loraga} while still applying metric amplitude control via $P_Q$ and Eq.~\eqref{eq:stage_objective}.
    \item \textbf{Full-Space Soft Reg:} Directly optimizes $\min_{\Delta W \in \mathbb{R}^{m\times d}} \mathcal{L}_t^{\mathrm{CE}}(W_{t-1} + \Delta W) + \frac{\lambda}{2}\operatorname{tr}(\Delta W M_t \Delta W^\top)$ over all $md$ parameters without a rank constraint.
    \item \textbf{PROUD-Full ($r=d$):} Our full-space whitened variant retains the stationary terminal interface and historical metric $M_t$, using the full input basis $P_t=L_t^{-\top}$. It parameterizes $\Delta W=ZL_t^{-1}$ with penalty $\frac{\lambda}{2}\|Z\|_F^2$, without low-rank truncation.
\end{enumerate}

\subsection{Continual-stage procedure}
\label{app:algorithm}
Algorithm~\ref{alg:proud} summarizes the stage-wise procedure from Section~\ref{sec:methodology}. The default experiments use $\lambda=0.01$ at every stage.
\begin{algorithm}[htbp]
\caption{PROUD at Continual Stage $t$}
\label{alg:proud}
\textbf{Input:} Reference weight $W_{t-1}$, historical second moment $C_{<t}$, dataset $\mathcal{D}_t$, preview subset $\mathcal{B}_t \subset \mathcal{D}_t$, rank $r$, damping $\rho$ (calibrated once on $\mathcal{B}_1$ before Stage~1), penalty $\lambda$. \\
\textbf{Output:} Updated weight $W_t$, updated second moment $C_{<t+1}$.
\begin{algorithmic}[1]
\raggedright
    \State \textbf{Preview Gradient:} Compute $\bar{G}_t = \nabla_W \mathcal{L}_t^{\mathrm{CE}}(W_{t-1}; \mathcal{B}_t)$ at fixed $W_{t-1}$ via Eq.~\eqref{eq:ce_loss}.
    \State \textbf{Metric Cholesky:} Form $M_t = C_{<t} + \rho I_d$ via Eq.~\eqref{eq:damped_metric} and factorize $M_t = L_t L_t^\top$.
    \State \textbf{Whitened SVD:} Solve $K_t L_t^\top = \bar{G}_t$ for $K_t$; extract top-$r$ right singular vectors $V_{t,r}$ of $K_t$ via SVD.
    \State \textbf{Basis Mapping:} Solve $L_t^\top P_t = V_{t,r}$ for $P_t \in \mathbb{R}^{d\times r}$ via Eq.~\eqref{eq:optimal_basis} (satisfying $P_t^\top M_t P_t = I_r$).
    \State \textbf{In-Stage Training:} Initialize $B_t \leftarrow \mathbf{0}_{m\times r}$. Freeze $W_{t-1}$ and $P_t$; optimize $B_t$ on $\mathcal{D}_t$ via Eq.~\eqref{eq:stage_objective}.
    \State \textbf{Stage Merge:} Update $W_t \leftarrow W_{t-1} + B_t P_t^\top$ and release temporary buffers $(\bar{G}_t, L_t, K_t, P_t, B_t)$.
    \State \textbf{Statistics Update:} Compute $\widehat{C}_t$ via Eq.~\eqref{eq:stage_cov} and update $C_{<t+1} \leftarrow \frac{(t-1)C_{<t} + \widehat{C}_t}{t}$.
\end{algorithmic}
\end{algorithm}

\FloatBarrier

\section{Details of continual evaluation metrics}
\label{app:metrics}

Let $\mathcal{A}_{s,t} \in [0, 100]$ denote the native evaluation score on the test set of task $s$ after completing training on stage $t \in \{1, \dots, T\}$. We report three standard continual learning metrics in the main text:
\begin{equation}
    \begin{aligned}
    \text{Last } (\bar{\mathcal{A}}) &= \frac{1}{T}\sum_{s=1}^T \mathcal{A}_{s,T}, \\
    \text{Avg} &= \frac{1}{T}\sum_{t=1}^T \left(\frac{1}{t}\sum_{s=1}^t \mathcal{A}_{s,t}\right), \\
    \text{BWT} &= \frac{1}{T-1}\sum_{s=1}^{T-1}\big(\mathcal{A}_{s,T} - \mathcal{A}_{s,s}\big).
    \end{aligned}
\end{equation}
Higher $\text{Last}$ and $\text{Avg}$ indicate stronger overall continual performance, while higher (less negative) $\text{BWT}$ indicates better knowledge retention.

\section{Detailed stage-by-stage result matrices}
\label{app:stage_matrices}

\begin{figure}[!htbp]
  \centering
  \includegraphics[width=\linewidth]{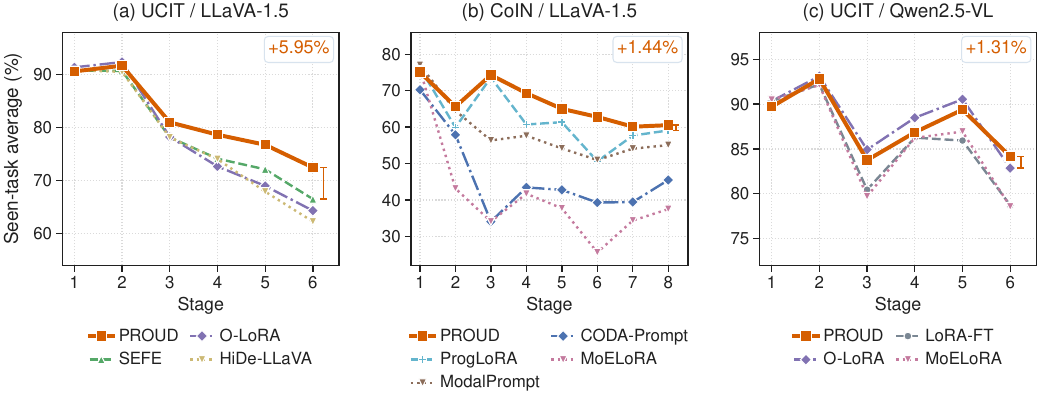}
  \caption{Complete stage curves for the compared methods. Each point is computed from the recorded result matrix and averages only the tasks seen so far; the evaluated set grows at each stage.}
  \label{fig:continual_full_app}
\end{figure}

Figure~\ref{fig:continual_full_app} plots the stage-wise seen-task accuracy curves, and Tables~\ref{tab:mat_ucit_llava}--\ref{tab:mat_replication_ucit} report the complete evaluation matrices $\mathcal{A}_{s,t}$ across all stages $t \in \{1, \dots, T\}$ on UCIT (LLaVA-v1.5-7B and Qwen2.5-VL-7B-Instruct) and CoIN (LLaVA-v1.5-7B). In each sub-table, row $t$ corresponds to the model evaluated immediately after training on stage $t$, column $s$ corresponds to the evaluated task ($s \le t$), bold diagonal entries denote initial accuracy $\mathcal{A}_{t,t}$, and the final column reports the average over all tasks seen up to stage $t$ (where the final row equals $\text{Last} / \bar{\mathcal{A}}$).

\begin{table*}[!htbp]
\centering
\proudtabsetup
\caption{Detailed stage-by-stage continual learning matrices $\mathcal{A}_{s,t}$ on \textbf{UCIT (LLaVA-v1.5-7B)} across representative methods and controlled subspace baselines ($r=128$). Tasks T1--T6 denote ImageNet-R, ArxivQA, VizWiz-caption, IconQA, CLEVR-Math, and Flickr30k.}
\label{tab:mat_ucit_llava}
\footnotesize
\setlength{\tabcolsep}{1.5pt}
\begin{tabularx}{\linewidth}{l*{7}{C{1}}@{\hspace{12pt}}l*{7}{C{1}}}
\toprule
\rowcolor{proudtableheadbg}
\textbf{Stage} & \textbf{T1} & \textbf{T2} & \textbf{T3} & \textbf{T4} & \textbf{T5} & \textbf{T6} & \textbf{Avg$_t$} &
\textbf{Stage} & \textbf{T1} & \textbf{T2} & \textbf{T3} & \textbf{T4} & \textbf{T5} & \textbf{T6} & \textbf{Avg$_t$} \\
\midrule
\multicolumn{8}{@{}>{\raggedright\arraybackslash}p{\dimexpr(\linewidth-12pt)/2\relax}@{\hspace{12pt}}}{\textbf{(a) PROUD (Default, seed 42)}\newline ($\text{Last}=72.49, \text{Avg}=81.89, \text{BWT}=-0.93$)} &
\multicolumn{8}{@{}>{\raggedright\arraybackslash}p{\dimexpr(\linewidth-12pt)/2\relax}@{}}{\textbf{(b) PROUD ($\gamma=0.03$, seed 42)}\newline ($\text{Last}=74.18, \text{Avg}=82.78, \text{BWT}=-1.23$)} \\
T1 & \textbf{90.63} & -- & -- & -- & -- & -- & 90.63 & T1 & \textbf{90.37} & -- & -- & -- & -- & -- & 90.37 \\
\rowcolor{proudtableshade}
T2 & 90.57 & \textbf{92.87} & -- & -- & -- & -- & 91.72 & T2 & 90.37 & \textbf{92.93} & -- & -- & -- & -- & 91.65 \\
T3 & 90.43 & 92.90 & \textbf{59.73} & -- & -- & -- & 81.02 & T3 & 90.13 & 92.90 & \textbf{61.73} & -- & -- & -- & 81.59 \\
\rowcolor{proudtableshade}
T4 & 90.23 & 92.80 & 62.01 & \textbf{69.70} & -- & -- & 78.69 & T4 & 90.03 & 92.63 & 62.65 & \textbf{76.00} & -- & -- & 80.33 \\
T5 & 90.20 & 92.93 & 62.59 & 69.53 & \textbf{68.67} & -- & 76.78 & T5 & 89.93 & 92.67 & 63.09 & 76.00 & \textbf{71.07} & -- & 78.55 \\
\rowcolor{proudtableshade}
T6 & 89.70 & 92.93 & 57.68 & 68.60 & 68.03 & \textbf{58.00} & \textbf{72.49} & T6 & 88.70 & 92.50 & 58.10 & 75.70 & 70.93 & \textbf{59.13} & \textbf{74.18} \\
\midrule
\multicolumn{8}{@{}>{\raggedright\arraybackslash}p{\dimexpr(\linewidth-12pt)/2\relax}@{\hspace{12pt}}}{\textbf{(c) PROUD (Default, seed 43)}\newline ($\text{Last}=72.81, \text{Avg}=82.01, \text{BWT}=-0.71$)} &
\multicolumn{8}{@{}>{\raggedright\arraybackslash}p{\dimexpr(\linewidth-12pt)/2\relax}@{}}{\textbf{(d) Seq.\ LoRA}\newline ($\text{Last}=47.63, \text{Avg}=56.01, \text{BWT}=-25.17$)} \\
T1 & \textbf{90.50} & -- & -- & -- & -- & -- & 90.50 & T1 & \textbf{72.83} & -- & -- & -- & -- & -- & 72.83 \\
\rowcolor{proudtableshade}
T2 & 90.63 & \textbf{92.67} & -- & -- & -- & -- & 91.65 & T2 & 35.87 & \textbf{94.03} & -- & -- & -- & -- & 64.95 \\
T3 & 90.27 & 92.70 & \textbf{60.38} & -- & -- & -- & 81.12 & T3 & 7.40 & 73.00 & \textbf{60.39} & -- & -- & -- & 46.93 \\
\rowcolor{proudtableshade}
T4 & 90.20 & 92.67 & 63.75 & \textbf{70.27} & -- & -- & 79.22 & T4 & 33.30 & 66.07 & 46.77 & \textbf{63.30} & -- & -- & 52.36 \\
T5 & 90.03 & 92.63 & 63.39 & 69.40 & \textbf{68.43} & -- & 76.78 & T5 & 28.43 & 68.17 & 43.39 & 55.33 & \textbf{61.63} & -- & 51.39 \\
\rowcolor{proudtableshade}
T6 & 89.13 & 92.67 & 59.96 & 68.73 & 68.23 & \textbf{58.14} & \textbf{72.81} & T6 & 29.63 & 76.63 & 44.65 & 42.77 & 32.63 & \textbf{59.45} & 47.63 \\
\midrule
\multicolumn{8}{@{}>{\raggedright\arraybackslash}p{\dimexpr(\linewidth-12pt)/2\relax}@{\hspace{12pt}}}{\textbf{(e) Historical Near-Null}\newline ($\text{Last}=61.01, \text{Avg}=74.78, \text{BWT}=-1.67$)} &
\multicolumn{8}{@{}>{\raggedright\arraybackslash}p{\dimexpr(\linewidth-12pt)/2\relax}@{}}{\textbf{(f) Current Gradient}\newline ($\text{Last}=67.89, \text{Avg}=80.09, \text{BWT}=-3.96$)} \\
T1 & \textbf{90.63} & -- & -- & -- & -- & -- & 90.63 & T1 & \textbf{90.63} & -- & -- & -- & -- & -- & 90.63 \\
\rowcolor{proudtableshade}
T2 & 90.37 & \textbf{90.17} & -- & -- & -- & -- & 90.27 & T2 & 90.63 & \textbf{91.87} & -- & -- & -- & -- & 91.25 \\
T3 & 90.47 & 89.30 & \textbf{45.46} & -- & -- & -- & 75.08 & T3 & 90.13 & 91.33 & \textbf{59.38} & -- & -- & -- & 80.28 \\
\rowcolor{proudtableshade}
T4 & 90.30 & 88.80 & 44.84 & \textbf{47.03} & -- & -- & 67.74 & T4 & 89.93 & 91.27 & 62.41 & \textbf{63.63} & -- & -- & 76.81 \\
T5 & 90.00 & 88.53 & 44.69 & 46.00 & \textbf{50.67} & -- & 63.98 & T5 & 89.37 & 91.10 & 61.65 & 62.40 & \textbf{63.93} & -- & 73.69 \\
\rowcolor{proudtableshade}
T6 & 89.37 & 88.47 & 41.70 & 45.70 & 50.37 & \textbf{50.44} & 61.01 & T6 & 86.77 & 91.50 & 48.59 & 59.90 & 62.90 & \textbf{57.69} & 67.89 \\
\midrule
\multicolumn{8}{@{}>{\raggedright\arraybackslash}p{\dimexpr(\linewidth-12pt)/2\relax}@{\hspace{12pt}}}{\textbf{(g) Random Basis}\newline ($\text{Last}=65.40, \text{Avg}=77.31, \text{BWT}=-7.31$)} &
\multicolumn{8}{@{}>{\raggedright\arraybackslash}p{\dimexpr(\linewidth-12pt)/2\relax}@{}}{\textbf{(h) Spectrum-Preserving Rotation}\newline ($\text{Last}=57.64, \text{Avg}=73.56, \text{BWT}=-23.54$)} \\
T1 & \textbf{88.53} & -- & -- & -- & -- & -- & 88.53 & T1 & \textbf{90.63} & -- & -- & -- & -- & -- & 90.63 \\
\rowcolor{proudtableshade}
T2 & 88.53 & \textbf{91.77} & -- & -- & -- & -- & 90.15 & T2 & 87.17 & \textbf{93.60} & -- & -- & -- & -- & 90.39 \\
T3 & 80.90 & 86.10 & \textbf{60.57} & -- & -- & -- & 75.86 & T3 & 53.77 & 91.67 & \textbf{63.57} & -- & -- & -- & 69.67 \\
\rowcolor{proudtableshade}
T4 & 81.43 & 87.97 & 57.85 & \textbf{65.87} & -- & -- & 73.28 & T4 & 60.40 & 84.23 & 48.97 & \textbf{82.17} & -- & -- & 68.94 \\
T5 & 82.70 & 87.27 & 60.28 & 59.10 & \textbf{63.80} & -- & 70.63 & T5 & 56.43 & 76.53 & 43.34 & 70.27 & \textbf{73.73} & -- & 64.06 \\
\rowcolor{proudtableshade}
T6 & 77.70 & 90.47 & 46.62 & 57.97 & 61.23 & \textbf{58.38} & 65.40 & T6 & 43.70 & 78.30 & 44.97 & 60.87 & 58.17 & \textbf{59.85} & 57.64 \\
\bottomrule
\end{tabularx}
\end{table*}

\begin{table*}[!htbp]
\centering
\proudtabsetup
\caption{Detailed stage-by-stage continual learning matrices $\mathcal{A}_{s,t}$ on \textbf{UCIT (Qwen2.5-VL-7B-Instruct)} across PROUD and multi-layer decoder baselines. Tasks T1--T6 denote ImageNet-R, ArxivQA, VizWiz-caption, IconQA, CLEVR-Math, and Flickr30k.}
\label{tab:mat_ucit_qwen}
\footnotesize
\setlength{\tabcolsep}{1.5pt}
\begin{tabularx}{\linewidth}{l*{7}{C{1}}@{\hspace{12pt}}l*{7}{C{1}}}
\toprule
\rowcolor{proudtableheadbg}
\textbf{Stage} & \textbf{T1} & \textbf{T2} & \textbf{T3} & \textbf{T4} & \textbf{T5} & \textbf{T6} & \textbf{Avg$_t$} &
\textbf{Stage} & \textbf{T1} & \textbf{T2} & \textbf{T3} & \textbf{T4} & \textbf{T5} & \textbf{T6} & \textbf{Avg$_t$} \\
\midrule
\multicolumn{8}{@{}>{\raggedright\arraybackslash}p{\dimexpr(\linewidth-12pt)/2\relax}@{\hspace{12pt}}}{\textbf{(a) PROUD (Ours, 0.46M params)}\newline ($\text{Last}=\mathbf{84.17}, \text{Avg}=87.78, \text{BWT}=-0.06$)} &
\multicolumn{8}{@{}>{\raggedright\arraybackslash}p{\dimexpr(\linewidth-12pt)/2\relax}@{}}{\textbf{(b) O-LoRA (Decoder, 15.14M/task)}\newline ($\text{Last}=82.86, \text{Avg}=88.37, \text{BWT}=-3.31$)} \\
T1 & \textbf{89.70} & -- & -- & -- & -- & -- & 89.70 & T1 & \textbf{90.33} & -- & -- & -- & -- & -- & 90.33 \\
\rowcolor{proudtableshade}
T2 & 90.00 & \textbf{95.63} & -- & -- & -- & -- & 92.82 & T2 & 90.37 & \textbf{95.87} & -- & -- & -- & -- & 93.12 \\
T3 & 89.63 & 95.73 & \textbf{65.89} & -- & -- & -- & 83.75 & T3 & 90.00 & 95.13 & \textbf{69.63} & -- & -- & -- & 84.92 \\
\rowcolor{proudtableshade}
T4 & 89.73 & 95.63 & 65.91 & \textbf{96.13} & -- & -- & 86.85 & T4 & 90.17 & 95.90 & 70.20 & \textbf{97.63} & -- & -- & 88.48 \\
T5 & 89.67 & 95.63 & 65.92 & 96.07 & \textbf{99.60} & -- & 89.38 & T5 & 90.00 & 95.87 & 69.95 & 97.20 & \textbf{99.67} & -- & 90.54 \\
\rowcolor{proudtableshade}
T6 & 89.37 & 95.73 & 65.87 & 96.07 & 99.60 & \textbf{58.35} & \textbf{84.17} & T6 & 90.03 & 95.90 & 54.14 & 97.10 & 99.40 & \textbf{60.58} & 82.86 \\
\midrule
\multicolumn{8}{@{}>{\raggedright\arraybackslash}p{\dimexpr(\linewidth-12pt)/2\relax}@{\hspace{12pt}}}{\textbf{(c) LoRA-FT (Decoder, 15.14M)}\newline ($\text{Last}=78.72, \text{Avg}=85.65, \text{BWT}=-8.59$)} &
\multicolumn{8}{@{}>{\raggedright\arraybackslash}p{\dimexpr(\linewidth-12pt)/2\relax}@{}}{\textbf{(d) MoELoRA (Decoder, 95.22M)}\newline ($\text{Last}=78.56, \text{Avg}=85.67, \text{BWT}=-8.80$)} \\
T1 & \textbf{90.33} & -- & -- & -- & -- & -- & 90.33 & T1 & \textbf{90.47} & -- & -- & -- & -- & -- & 90.47 \\
\rowcolor{proudtableshade}
T2 & 88.63 & \textbf{95.77} & -- & -- & -- & -- & 92.20 & T2 & 88.57 & \textbf{95.67} & -- & -- & -- & -- & 92.12 \\
T3 & 75.93 & 95.53 & \textbf{69.89} & -- & -- & -- & 80.45 & T3 & 74.43 & 95.20 & \textbf{69.50} & -- & -- & -- & 79.71 \\
\rowcolor{proudtableshade}
T4 & 80.87 & 95.50 & 70.00 & \textbf{98.67} & -- & -- & 86.26 & T4 & 81.93 & 95.50 & 68.59 & \textbf{98.90} & -- & -- & 86.23 \\
T5 & 69.07 & 95.53 & 67.45 & 98.00 & \textbf{99.63} & -- & 85.94 & T5 & 73.23 & 95.80 & 67.65 & 98.20 & \textbf{99.80} & -- & 86.94 \\
\rowcolor{proudtableshade}
T6 & 66.67 & 95.60 & 51.75 & 98.17 & 99.17 & \textbf{60.96} & 78.72 & T6 & 65.97 & 95.53 & 51.19 & 98.13 & 99.50 & \textbf{61.05} & 78.56 \\
\bottomrule
\end{tabularx}
\end{table*}

\begin{figure}[p]
  \centering
  \includegraphics[width=\linewidth]{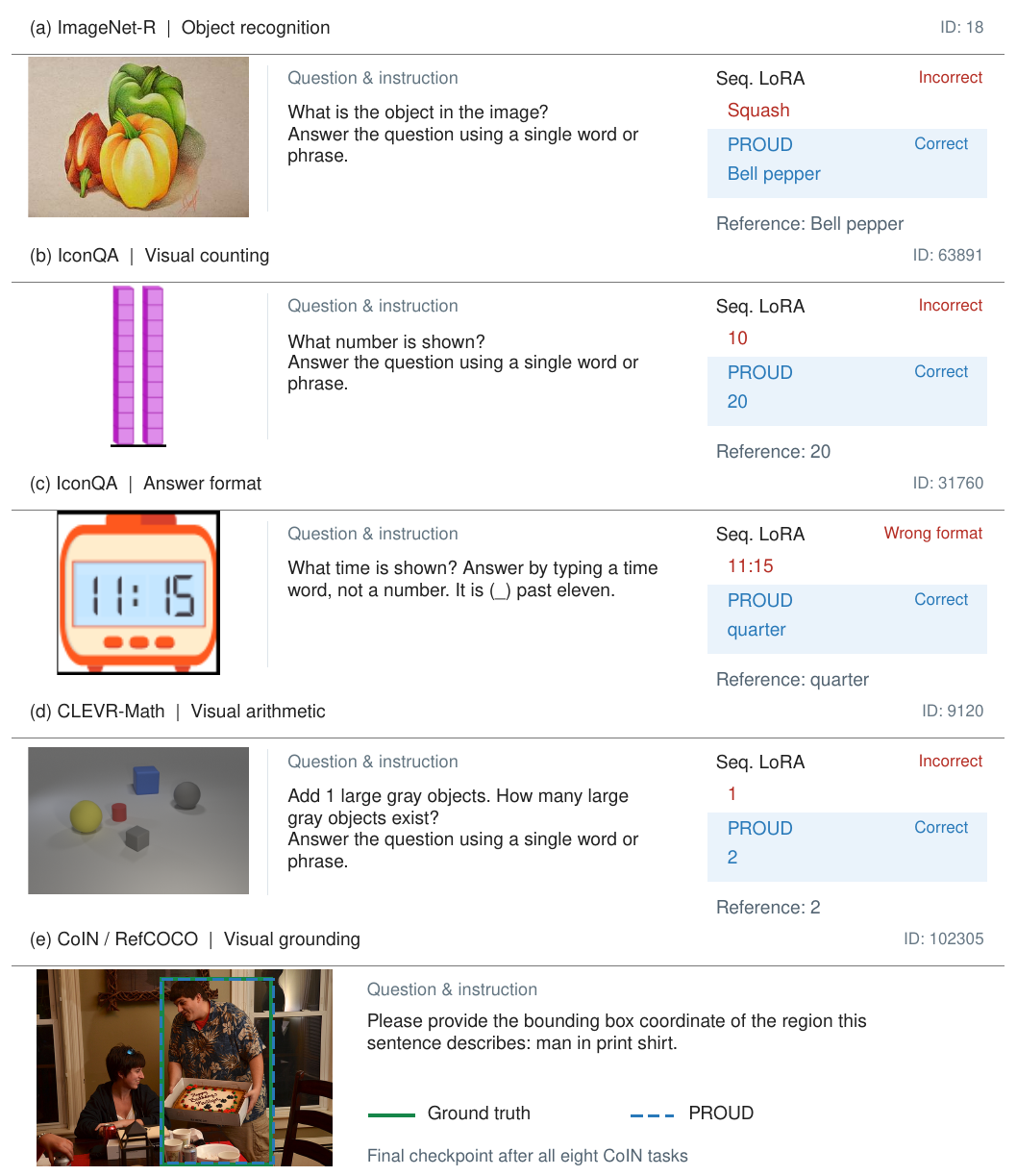}
  \caption{\textbf{Selected qualitative examples on UCIT and CoIN.} Panels (a)--(d) compare PROUD with Seq.\ LoRA after all six UCIT tasks at the terminal projector layer of LLaVA-v1.5-7B ($r=128$), using original images, full prompts, and verbatim predictions. Panels (a), (b), and (d) illustrate recognition, counting, and arithmetic errors; (c) shows a correct time in the wrong requested format. Panel (e) shows a RefCOCO example after all eight CoIN tasks, with the ground-truth box in green and PROUD's recorded prediction in dashed blue. These selected examples complement the aggregate results.}
  \label{fig:qualitative}
\end{figure}

\begin{table*}[!htbp]
\centering
\proudtabsetup
\caption{Detailed stage-by-stage matrices on \textbf{CoIN (LLaVA-v1.5-7B)} under the default uniform penalty $\lambda=0.01$ and high-exposure Stage-7 regularization $\lambda_7=0.3$, as detailed in Section~\ref{app:ablations}. Both use learning rate $8\times10^{-5}$ at every stage and seed 42. The control changes only the VQAv2 penalty from $\lambda_7=0.01$ to $0.3$.}
\label{tab:mat_coin_llava}
\footnotesize
\setlength{\tabcolsep}{2pt}
\begin{tabularx}{\linewidth}{>{\raggedright\arraybackslash}p{0.19\linewidth}*{9}{C{1}}}
\toprule
\rowcolor{proudtableheadbg}
\textbf{Stage Trained} & \textbf{SciQA} & \textbf{TextVQA} & \textbf{ImgNet} & \textbf{GQA} & \textbf{VizWiz} & \textbf{Ground.} & \textbf{VQAv2} & \textbf{OCRVQA} & \proudtablehead{Seen\\Avg$_t$} \\
\midrule
\multicolumn{10}{@{}>{\raggedright\arraybackslash}p{\linewidth}@{}}{\textbf{(a) PROUD (Uniform Default)}\newline ($\text{Last}=60.53,\ \text{Avg}=66.59,\ \text{BWT}=-3.66$)} \\
1: ScienceQA & \textbf{75.10} & -- & -- & -- & -- & -- & -- & -- & 75.10 \\
\rowcolor{proudtableshade}
2: TextVQA & 75.31 & \textbf{55.91} & -- & -- & -- & -- & -- & -- & 65.61 \\
3: ImageNet & 75.05 & 53.70 & \textbf{94.16} & -- & -- & -- & -- & -- & 74.30 \\
\rowcolor{proudtableshade}
4: GQA & 75.64 & 55.27 & 88.55 & \textbf{57.75} & -- & -- & -- & -- & 69.30 \\
5: VizWiz & 75.62 & 54.20 & 87.25 & 57.66 & \textbf{50.31} & -- & -- & -- & 65.01 \\
\rowcolor{proudtableshade}
6: Grounding & 75.62 & 54.68 & 86.53 & 57.58 & 49.64 & \textbf{52.72} & -- & -- & 62.80 \\
7: VQAv2 & 75.48 & 55.44 & 67.15 & 57.07 & 51.86 & 51.25 & \textbf{62.49} & -- & 60.11 \\
\rowcolor{proudtableresult}
8: OCR-VQA (Final) & 75.67 & 56.18 & 66.46 & 57.36 & 52.56 & 51.92 & 62.68 & \textbf{61.41} & 60.53 \\
\midrule
\multicolumn{10}{@{}>{\raggedright\arraybackslash}p{\linewidth}@{}}{\textbf{(b) PROUD (Stage-7 regularization control)}\newline ($\text{Last}=61.36,\ \text{Avg}=66.81,\ \text{BWT}=-2.33$)} \\
1--6: (Identical prefix) & 75.62 & 54.68 & 86.53 & 57.58 & 49.64 & \textbf{52.72} & -- & -- & 62.80 \\
\rowcolor{proudtableshade}
7: VQAv2 & 75.48 & 54.82 & 77.90 & 56.85 & 50.03 & 52.06 & \textbf{60.04} & -- & 61.03 \\
\rowcolor{proudtableresult}
8: OCR-VQA (Final) & 75.90 & 55.85 & 77.49 & 56.72 & 51.24 & 52.30 & 60.20 & \textbf{61.20} & 61.36 \\
\bottomrule
\end{tabularx}
\end{table*}
\FloatBarrier
\begin{samepage}
\noindent\textbf{Answer-format retention.}
Updating multi-layer LLM decoders can overwrite answer-format conventions: for example, lowercase VQA annotations can interfere with uppercase multiple-choice outputs on ScienceQA. Freezing the decoder and calibrating only the projector bridge keeps ScienceQA stable at $75.10\% \rightarrow 75.67\%$ across all eight CoIN stages without drop--rebound oscillations (Table~\ref{tab:mat_coin_llava}). Figure~\ref{fig:qualitative} provides complementary examples of visual counting and instruction-format preservation at the final checkpoint, including a case where sequential LoRA predicts the correct time in the wrong requested format. Panel (e) adds a CoIN RefCOCO localization example with the recorded final-stage PROUD prediction and the dataset bounding box.

We also audit all 48 questions in the 3,000-example IconQA evaluation split that explicitly request a ``time word, not a number,'' using the saved final-stage predictions. A digit-format violation is any decimal digit in the generated answer. Seq.\ LoRA violates this instruction on 43/48 questions (89.6\%), compared with 0/48 (0.0\%) for PROUD. These violations account for 43 of Seq.\ LoRA's 45 incorrect answers (95.6\%) within this subset; PROUD answers all 48 correctly. These rates apply to the explicit time-word subset, not the entire IconQA evaluation set.
\end{samepage}

\begin{table*}[!htbp]
\centering
\proudtabsetup
\caption{Additional complete UCIT stage matrices for seed replication, small preview budgets, and metric controls. Task order follows Table~\ref{tab:mat_ucit_llava}. Preview and metric controls use seed 42. The original default runs are reported in Tables~\ref{tab:mat_ucit_llava} and~\ref{tab:mat_ucit_qwen}.}
\label{tab:mat_replication_ucit}
\footnotesize
\setlength{\tabcolsep}{1.5pt}
\begin{tabularx}{\linewidth}{l*{7}{C{1}}@{\hspace{12pt}}l*{7}{C{1}}}
\toprule
\rowcolor{proudtableheadbg}
\textbf{Stage} & \textbf{T1} & \textbf{T2} & \textbf{T3} & \textbf{T4} & \textbf{T5} & \textbf{T6} & \textbf{Avg$_t$} & \textbf{Stage} & \textbf{T1} & \textbf{T2} & \textbf{T3} & \textbf{T4} & \textbf{T5} & \textbf{T6} & \textbf{Avg$_t$} \\
\midrule
\multicolumn{8}{@{}>{\raggedright\arraybackslash}p{\dimexpr(\linewidth-12pt)/2\relax}@{\hspace{12pt}}}{\textbf{(a) PROUD / LLaVA, seed 44}\newline ($\text{Last}=72.29,\ \text{Avg}=81.49,\ \text{BWT}=-0.81$)} & \multicolumn{8}{@{}>{\raggedright\arraybackslash}p{\dimexpr(\linewidth-12pt)/2\relax}@{}}{\textbf{(b) PROUD-Full ($r=d$) / LLaVA, seed 42}\newline ($\text{Last}=72.28,\ \text{Avg}=82.40,\ \text{BWT}=-0.84$)} \\
T1 & \textbf{90.43} & -- & -- & -- & -- & -- & 90.43 & T1 & \textbf{92.07} & -- & -- & -- & -- & -- & 92.07 \\
\rowcolor{proudtableshade}
T2 & 90.53 & \textbf{92.80} & -- & -- & -- & -- & 91.66 & T2 & 92.13 & \textbf{92.97} & -- & -- & -- & -- & 92.55 \\
T3 & 90.20 & 92.73 & \textbf{58.85} & -- & -- & -- & 80.59 & T3 & 92.00 & 93.07 & \textbf{55.57} & -- & -- & -- & 80.21 \\
\rowcolor{proudtableshade}
T4 & 90.07 & 92.60 & 60.02 & \textbf{69.17} & -- & -- & 77.97 & T4 & 92.07 & 93.10 & 59.66 & \textbf{74.90} & -- & -- & 79.93 \\
T5 & 90.23 & 92.60 & 59.47 & 69.23 & \textbf{68.47} & -- & 76.00 & T5 & 92.03 & 93.13 & 60.19 & 74.33 & \textbf{67.20} & -- & 77.38 \\
\rowcolor{proudtableshade}
T6 & 89.07 & 92.47 & 56.17 & 69.77 & 68.20 & \textbf{58.06} & \textbf{72.29} & T6 & 91.07 & 93.20 & 53.38 & 73.93 & 66.93 & \textbf{55.17} & \textbf{72.28} \\
\midrule
\multicolumn{8}{@{}>{\raggedright\arraybackslash}p{\dimexpr(\linewidth-12pt)/2\relax}@{\hspace{12pt}}}{\textbf{(c) PROUD / Qwen, seed 43}\newline ($\text{Last}=84.53,\ \text{Avg}=87.88,\ \text{BWT}=0.17$)} & \multicolumn{8}{@{}>{\raggedright\arraybackslash}p{\dimexpr(\linewidth-12pt)/2\relax}@{}}{\textbf{(d) PROUD / Qwen, seed 44}\newline ($\text{Last}=84.09,\ \text{Avg}=87.74,\ \text{BWT}=-0.16$)} \\
T1 & \textbf{89.13} & -- & -- & -- & -- & -- & 89.13 & T1 & \textbf{89.50} & -- & -- & -- & -- & -- & 89.50 \\
\rowcolor{proudtableshade}
T2 & 89.23 & \textbf{95.57} & -- & -- & -- & -- & 92.40 & T2 & 89.63 & \textbf{95.33} & -- & -- & -- & -- & 92.48 \\
T3 & 89.17 & 95.57 & \textbf{67.60} & -- & -- & -- & 84.11 & T3 & 89.50 & 95.37 & \textbf{66.58} & -- & -- & -- & 83.82 \\
\rowcolor{proudtableshade}
T4 & 89.13 & 95.63 & 68.19 & \textbf{96.50} & -- & -- & 87.36 & T4 & 89.47 & 95.40 & 66.75 & \textbf{96.43} & -- & -- & 87.01 \\
T5 & 89.20 & 95.70 & 67.65 & 96.77 & \textbf{99.47} & -- & 89.76 & T5 & 89.50 & 95.40 & 66.70 & 96.47 & \textbf{99.50} & -- & 89.51 \\
\rowcolor{proudtableshade}
T6 & 89.17 & 95.63 & 68.17 & 96.63 & 99.50 & \textbf{58.08} & \textbf{84.53} & T6 & 88.70 & 95.30 & 66.65 & 96.40 & 99.50 & \textbf{58.01} & \textbf{84.09} \\
\midrule
\multicolumn{8}{@{}>{\raggedright\arraybackslash}p{\dimexpr(\linewidth-12pt)/2\relax}@{\hspace{12pt}}}{\textbf{(e) PROUD / LLaVA, preview 32}\newline ($\text{Last}=71.57,\ \text{Avg}=80.79,\ \text{BWT}=-0.47$)} & \multicolumn{8}{@{}>{\raggedright\arraybackslash}p{\dimexpr(\linewidth-12pt)/2\relax}@{}}{\textbf{(f) PROUD / LLaVA, preview 64}\newline ($\text{Last}=72.54,\ \text{Avg}=81.42,\ \text{BWT}=-0.22$)} \\
T1 & \textbf{89.80} & -- & -- & -- & -- & -- & 89.80 & T1 & \textbf{90.43} & -- & -- & -- & -- & -- & 90.43 \\
\rowcolor{proudtableshade}
T2 & 89.93 & \textbf{92.83} & -- & -- & -- & -- & 91.38 & T2 & 90.47 & \textbf{93.10} & -- & -- & -- & -- & 91.79 \\
T3 & 89.70 & 92.80 & \textbf{57.76} & -- & -- & -- & 80.09 & T3 & 90.13 & 93.10 & \textbf{56.77} & -- & -- & -- & 80.00 \\
\rowcolor{proudtableshade}
T4 & 89.60 & 92.77 & 58.69 & \textbf{66.57} & -- & -- & 76.91 & T4 & 90.17 & 93.03 & 57.91 & \textbf{71.23} & -- & -- & 78.09 \\
T5 & 89.70 & 92.77 & 58.38 & 66.60 & \textbf{67.47} & -- & 74.98 & T5 & 90.23 & 92.87 & 57.73 & 70.57 & \textbf{67.00} & -- & 75.68 \\
\rowcolor{proudtableshade}
T6 & 88.80 & 92.77 & 57.31 & 65.60 & 67.60 & \textbf{57.36} & \textbf{71.57} & T6 & 89.73 & 93.00 & 57.18 & 70.30 & 67.23 & \textbf{57.78} & \textbf{72.54} \\
\midrule
\multicolumn{8}{@{}>{\raggedright\arraybackslash}p{\dimexpr(\linewidth-12pt)/2\relax}@{\hspace{12pt}}}{\textbf{(g) PROUD / LLaVA, preview 128}\newline ($\text{Last}=72.41,\ \text{Avg}=81.63,\ \text{BWT}=-0.64$)} & \multicolumn{8}{@{}>{\raggedright\arraybackslash}p{\dimexpr(\linewidth-12pt)/2\relax}@{}}{\textbf{(h) Input--output sensitivity / LLaVA}\newline ($\text{Last}=72.38,\ \text{Avg}=81.65,\ \text{BWT}=-0.75$)} \\
T1 & \textbf{90.50} & -- & -- & -- & -- & -- & 90.50 & T1 & \textbf{90.63} & -- & -- & -- & -- & -- & 90.63 \\
\rowcolor{proudtableshade}
T2 & 90.70 & \textbf{92.67} & -- & -- & -- & -- & 91.69 & T2 & 90.53 & \textbf{92.93} & -- & -- & -- & -- & 91.73 \\
T3 & 90.33 & 92.63 & \textbf{59.25} & -- & -- & -- & 80.74 & T3 & 90.47 & 92.87 & \textbf{60.37} & -- & -- & -- & 81.24 \\
\rowcolor{proudtableshade}
T4 & 90.30 & 92.73 & 60.40 & \textbf{69.33} & -- & -- & 78.19 & T4 & 90.43 & 92.83 & 60.99 & \textbf{67.70} & -- & -- & 77.99 \\
T5 & 90.37 & 92.70 & 60.66 & 69.37 & \textbf{68.13} & -- & 76.25 & T5 & 90.47 & 92.83 & 60.89 & 67.57 & \textbf{67.87} & -- & 75.93 \\
\rowcolor{proudtableshade}
T6 & 90.03 & 92.63 & 58.03 & 68.43 & 67.57 & \textbf{57.77} & \textbf{72.41} & T6 & 89.80 & 92.83 & 57.98 & 67.07 & 68.07 & \textbf{58.51} & \textbf{72.38} \\
\bottomrule
\end{tabularx}
\end{table*}

\FloatBarrier
\section{Extended experimental results and ablations}
\label{app:ablations}


\begin{table*}[!htbp]
\centering
\proudtabsetup
\caption{Equal-budget local probes ($\varepsilon=0.1$, $r=128$) and 300-step development trajectories from a shared Stage-2 checkpoint to two unseen target tasks (\textbf{VizWiz-caption} and \textbf{IconQA}). All bases satisfy $P_Q^\top M_3 P_Q=I_r$ and incur the exact same metric cost $q_3(\Delta W)=\varepsilon^2$. Probe capture is computed from the mean loss gradient on the 128-example held-out split $\mathcal{D}_{\mathrm{eval}}$, disjoint from basis selection ($\mathcal{D}_{\mathrm{sel}}$) and step estimation ($\mathcal{D}_{\mathrm{step}}$). The trajectories in Table~\ref{tab:app_trajectory} use a separate 32-example diagnostic split. Step-300 val CE denotes the held-out cross-entropy loss after 300 optimization steps (4,800 examples). The shared Stage-2 checkpoint uses rank 32; candidate updates use rank 128. The matched budget applies to $q_3$, while independently measured historical drift may differ.}
\label{tab:probing}
\footnotesize
\setlength{\tabcolsep}{2pt}
\begin{tabularx}{\linewidth}{l*{4}{C{1}}*{4}{C{1}}}
\toprule
\rowcolor{proudtableheadbg}
& \multicolumn{4}{c}{\textbf{Target Task: VizWiz-caption}} & \multicolumn{4}{c}{\textbf{Target Task: IconQA}} \\
\cmidrule(lr){2-5} \cmidrule(lr){6-9}
\rowcolor{proudtableheadbg}
\textbf{Input Subspace} & \proudtablehead{Probe grad\\capture $\uparrow$} & \proudtablehead{Probe CE\\drop $\uparrow$} & \proudtablehead{Gain / drift\\ratio $\uparrow$} & \proudtablehead{Step-300\\val CE $\downarrow$} & \proudtablehead{Probe grad\\capture $\uparrow$} & \proudtablehead{Probe CE\\drop $\uparrow$} & \proudtablehead{Gain / drift\\ratio $\uparrow$} & \proudtablehead{Step-300\\val CE $\downarrow$} \\
\midrule
Random & 4.42\% & 0.0417 & 4.30 & 2.6584 & 3.77\% & 0.1064 & 10.79 & 0.5127 \\
\rowcolor{proudtableshade}
Historical Near-Null & 3.47\% & 0.0143 & 1.16 & 3.3139 & 3.45\% & 0.0904 & 7.30 & 0.5996 \\
Current Gradient & 11.54\% & 0.0722 & 6.96 & 2.4890 & 19.89\% & 0.2137 & 20.43 & 0.4787 \\
\midrule
\rowcolor{proudtableresult}
\textbf{PROUD (Ours)} & \textbf{39.77\%} & \textbf{0.1346} & \textbf{9.25} & \textbf{2.1730} & \textbf{86.48\%} & \textbf{0.3070} & \textbf{23.26} & \textbf{0.4636} \\
\bottomrule
\end{tabularx}
\end{table*}

\begin{table*}[!htbp]
\centering
\proudtabsetup
\caption{Sensitivity and robustness analysis of PROUD on UCIT across rank $r$, damping $\gamma$, sample counts, and task orderings. Rank sweep is evaluated under $\mathrm{lr}=2\times 10^{-4}$; default $\mathrm{lr}=8\times 10^{-5}$ yields 72.49\% at $r=128$.}
\label{tab:sensitivity}
\footnotesize
\setlength{\tabcolsep}{3pt}
\begin{tabularx}{\linewidth}{lC{1}C{1}lC{1}C{1}}
\toprule
\rowcolor{proudtableheadbg}
\textbf{Setting} & \textbf{Last} $\uparrow$ & \textbf{BWT} $\uparrow$ & \textbf{Setting} & \textbf{Last} $\uparrow$ & \textbf{BWT} $\uparrow$ \\
\midrule
\multicolumn{3}{@{}l}{\textit{Rank $r$ ($\mathrm{lr}=2\times 10^{-4}$)}} & \multicolumn{3}{l}{\textit{Preview / History samples}} \\
$r=32$ & 68.87 & -0.42 & Prev 256 / Hist 512 & 72.49 & -0.93 \\
\rowcolor{proudtableshade}
$r=64$ & 69.74 & -1.22 & Prev 512 / Hist 512 & 72.54 & -0.62 \\
$r=96$ & 70.89 & -0.13 & Prev 1024 / Hist 512 & \textbf{72.96} & -0.79 \\
$r=128$ & 71.07 & -0.59 & Prev 256 / Hist 1024 & 72.02 & -1.31 \\
\rowcolor{proudtableshade}
$r=160$ & \textbf{71.11} & -0.17 & Prev 256 / Hist 2048 & 71.90 & -1.38 \\
$r=256$ & 70.21 & -1.13 & Prev 256 / Hist All & 72.04 & -1.23 \\
\midrule
\multicolumn{3}{@{}l}{\textit{Damping $\gamma$ ($r=128$)}} & \multicolumn{3}{l}{\textit{Task sequence ordering}} \\
$\gamma = 0.003$ & 69.64 & \textbf{-0.66} & Standard order & \textbf{72.49} & \textbf{-0.93} \\
\rowcolor{proudtableshade}
$\gamma = 0.01$ & 72.49 & -0.93 & Reverse order & 71.33 & -2.09 \\
$\gamma = 0.03$ & \textbf{74.18} & -1.23 & Random order & 71.02 & -1.30 \\
\bottomrule
\end{tabularx}
\end{table*}

\begin{table*}[!htbp]
\centering
\proudtabsetup
\caption{Controlled continual learning comparisons at the shared terminal projector layer on UCIT (LLaVA-v1.5-7B, $T=6$). All low-rank methods use $r=128$ and the exact same training budget.}
\label{tab:main_cl_detailed}
\footnotesize
\setlength{\tabcolsep}{2pt}
\begin{tabularx}{\linewidth}{>{\raggedright\arraybackslash}p{0.20\linewidth}C{1.10}C{1.12}C{1.06}C{0.95}C{0.95}C{0.92}C{1.16}C{1.04}C{0.75}C{0.95}}
\toprule
\rowcolor{proudtableheadbg}
\textbf{Method / Subspace} & \textbf{Params} & \textbf{ImgNet-R} & \textbf{ArxivQA} & \textbf{VizWiz} & \textbf{IconQA} & \textbf{CLEVR} & \textbf{Flickr30k} & \proudtablehead{\textbf{Last} $\uparrow$} & \proudtablehead{\textbf{Avg} $\uparrow$} & \proudtablehead{\textbf{BWT} $\uparrow$} \\
\midrule
Zero-shot & 0 & 16.17 & 54.20 & 41.59 & 20.30 & 20.33 & 45.62 & 33.04 & -- & -- \\
\rowcolor{proudtableshade}
Seq.\ Full-Rank FT & 16.78M & 37.70 & 77.63 & 45.73 & 62.37 & 60.17 & \textbf{61.29} & 57.48 & 67.84 & -24.80 \\
Seq.\ LoRA & 1.05M & 29.63 & 76.63 & 44.65 & 42.77 & 32.63 & 59.45 & 47.63 & 56.01 & -25.17 \\
Full-Space Soft Reg & 16.78M & 73.30 & 91.70 & 48.87 & \textbf{79.57} & \textbf{73.67} & 60.52 & 71.27 & 80.49 & -7.60 \\
\rowcolor{proudtableshade}
PROUD-Full ($r=d$) & 16.78M & \textbf{91.07} & \textbf{93.20} & 53.38 & 73.93 & 66.93 & 55.17 & 72.28 & \textbf{82.40} & \textbf{-0.84} \\
\midrule
\multicolumn{11}{l}{\textit{Fixed input subspaces with matched rank ($r=128$) and isometric displacement regularization}} \\
Random Basis & 0.52M & 77.70 & 90.47 & 46.62 & 57.97 & 61.23 & 58.38 & 65.40 & 77.31 & -7.31 \\
\rowcolor{proudtableshade}
Historical Near-Null~\cite{liang2024inflora,saha2021gpm} & 0.52M & 89.37 & 88.47 & 41.70 & 45.70 & 50.37 & 50.44 & 61.01 & 74.78 & -1.67 \\
Current Gradient~\cite{wang2024loraga} & 0.52M & 86.77 & 91.50 & 48.59 & 59.90 & 62.90 & 57.69 & 67.89 & 80.09 & -3.96 \\
\midrule
\rowcolor{proudtableresult}
\textbf{PROUD (Ours)} & \textbf{0.52M} & 89.70 & 92.93 & \textbf{57.68} & 68.60 & 68.03 & 58.00 & \textbf{72.49} & 81.89 & -0.93 \\
\bottomrule
\end{tabularx}
\end{table*}

\begin{table*}[!htbp]
\centering
\proudtabsetup
\caption{Historical metric and adaptation-locus results on UCIT with 524,288 trainable parameters. CLIP layers use $r=512$; the remaining sites use $r=128$.}
\label{tab:ablations_detailed}
\footnotesize
\setlength{\tabcolsep}{3pt}
\begin{tabularx}{\linewidth}{>{\raggedright\arraybackslash}p{0.66\linewidth}*{2}{C{1}}}
\toprule
\rowcolor{proudtableheadbg}
\textbf{Variant / Configuration} & \proudtablehead{\textbf{Last} $\uparrow$} & \proudtablehead{\textbf{BWT} $\uparrow$} \\
\midrule
\rowcolor{proudtableresult}
\textbf{PROUD (Default, Projector Final Layer)} & \textbf{72.49} & \textbf{-0.93} \\
\midrule
\multicolumn{3}{@{}l}{\textit{Historical metric ($M_t$) ablations}} \\
w/o Historical metric ($C_{<t}=\mathbf{0}$) & 53.39 & -28.13 \\
\rowcolor{proudtableshade}
w/o Spectral orientation (Random rotation) & 57.64 & -23.54 \\
w/o Stage accumulation ($\widehat{C}_{t-1}$ only) & 68.40 & -7.64 \\
\midrule
\multicolumn{3}{@{}l}{\textit{Adaptation locus comparison}} \\
\rowcolor{proudtableshade}
Projector First Linear Layer & 66.80 & -3.16 \\
LLM Decoder Shallow (Layer 0, V-proj) & 32.43 & -22.52 \\
\rowcolor{proudtableshade}
LLM Decoder Middle (Layer 15, V-proj) & 64.22 & -6.14 \\
LLM Decoder Deep (Layer 31, V-proj) & 52.38 & -3.79 \\
\rowcolor{proudtableshade}
CLIP Vision Shallow (Layer 0, V-proj) & 32.89 & -0.06 \\
CLIP Vision Middle (Layer 11, V-proj) & 46.64 & -1.94 \\
CLIP Vision Deep (Layer 22, V-proj) & 51.11 & -4.01 \\
\bottomrule
\end{tabularx}
\end{table*}

\begin{figure}[!tbp]
  \centering
  \includegraphics[width=\linewidth]{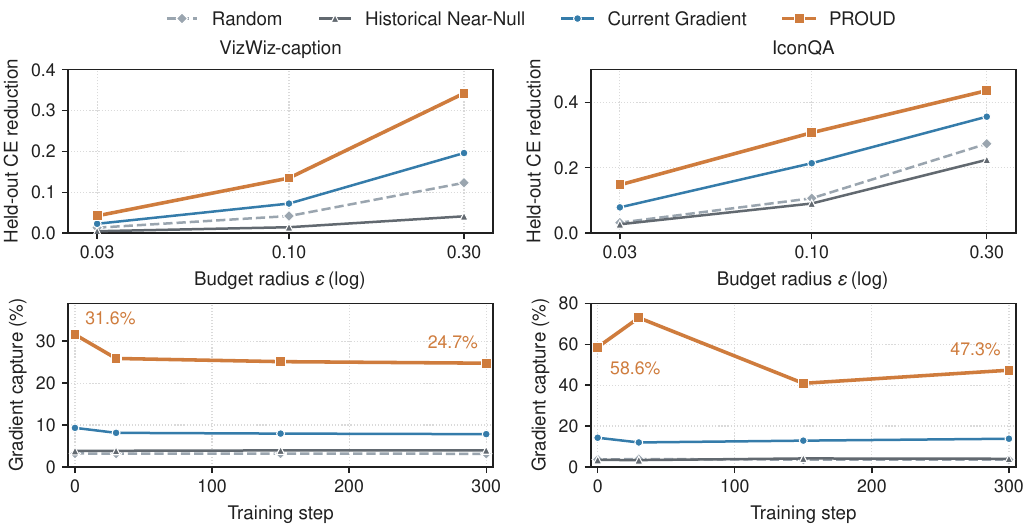}
  \caption{Equal-budget local probes (logarithmic budget axes) and 300-step fixed-basis training trajectories on VizWiz-caption and IconQA, branching from a shared Stage-2 checkpoint under matched metric cost $q_3(\Delta W)=\varepsilon^2$ ($r=128$). Probe evaluation uses 128 held-out examples; trajectory capture uses mean loss gradients on a separate fixed 32-example diagnostic split.}
  \label{fig:probes_app}
\end{figure}

\begin{figure}[!tbp]
  \centering
  \includegraphics[width=\linewidth]{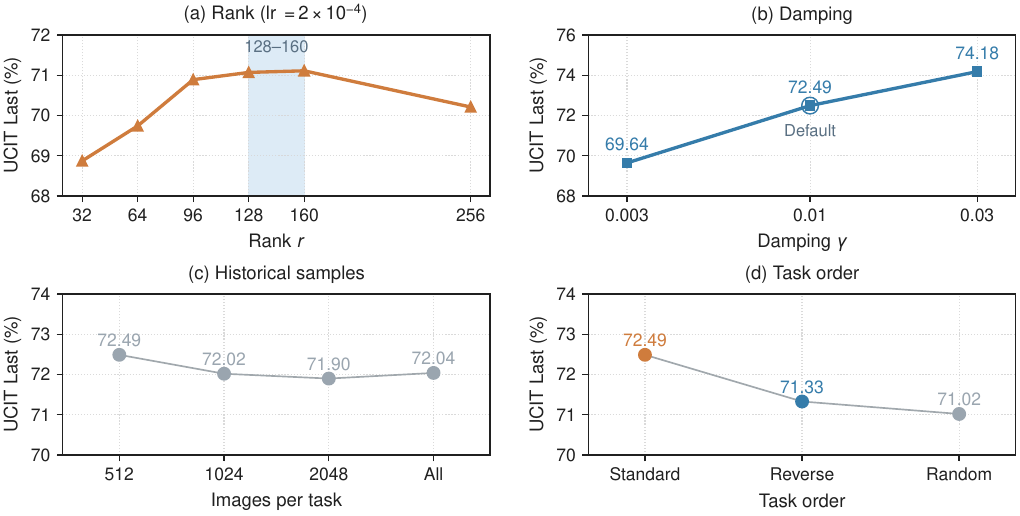}
  \caption{Sensitivity and robustness on UCIT. The rank sweep uses learning rate $2\times10^{-4}$; other panels use $8\times10^{-5}$ and rank 128. Damping is displayed on a logarithmic horizontal axis. History counts and task orders are categorical settings. Table~\ref{tab:sensitivity} retains the numerical results.}
  \label{fig:sensitivity_app}
\end{figure}

\noindent\textbf{Equal-budget probe construction.}
The Stage~$2\to3$ diagnostic uses disjoint $\mathcal{D}_{\mathrm{sel}}$ for basis extraction, $\mathcal{D}_{\mathrm{step}}$ for step estimation, and $\mathcal{D}_{\mathrm{eval}}$ for held-out evaluation. For a metric-orthonormal basis $P_Q$, the step is $\Delta W_{\mathrm{probe}}(P_Q)=-\varepsilon\frac{\bar{G}_{\mathrm{step}}P_Q}{\|\bar{G}_{\mathrm{step}}P_Q\|_F}P_Q^\top$, giving $q_3(\Delta W_{\mathrm{probe}})=\varepsilon^2$ for every candidate subspace. Table~\ref{tab:probing} and Figure~\ref{fig:probes_app} report the independent evaluation results and 300-step trajectories. The single-step probes use a 128-example evaluation split; the training trajectories use a separate fixed 32-example diagnostic split, as detailed below Table~\ref{tab:app_trajectory}.

\noindent\textbf{Interpreting the controlled subspaces.}
Table~\ref{tab:main_cl_detailed} compares unconstrained adaptation (Seq.\ Full-Rank FT and Seq.\ LoRA), two protected full-space references, and rank-matched fixed subspaces at the same terminal projector layer; their formulations are given in Section~\ref{app:controlled_baselines}. Historical Near-Null limits forgetting ($\text{BWT}=-1.67$) but severely restricts plasticity (61.01\% Last, vs.\ 65.40\% for Random). Current Gradient aligns with $\bar{G}_t$ but overlaps with high-energy historical directions where regularization resists movement (67.89\% Last). By jointly ranking benefit against historical cost, PROUD achieves 72.49\% Last and $-0.93$ BWT, surpassing Full-Space Soft Reg (71.27\%). Although Full-Space Soft Reg and PROUD-Full ($r=d$) have equivalent objectives under $\Delta W=ZL_t^{-1}$, Adam's element-wise moment normalization~\citep{kingma2015adam} is not invariant to this general linear reparameterization, so their finite-step optimization trajectories need not coincide. In PROUD-Full, the gradient is transformed by $L_t^{-\top}$ and a coordinate update $\delta Z$ maps back as $\delta\Delta W=\delta ZL_t^{-1}$, making the trajectory depend on the historical metric. The reported runs yield $-7.60\%$ BWT for Full-Space Soft Reg and $-0.84\%$ BWT with 72.28\% Last for PROUD-Full. Rank truncation to $r=128$ further filters out low-benefit directions and improves Last to 72.49\% with $32\times$ fewer trainable parameters. On Qwen2.5-VL-7B, rank truncation similarly improves continual performance over PROUD-Full ($r=d$, 18.35M parameters), raising Last from $82.85\%$ to $84.17\%$ and Avg from $87.18\%$ to $87.78\%$ with $40\times$ fewer trainable parameters while maintaining near-zero forgetting ($-0.06\%$ vs.\ $+0.17\%$ BWT). The complete PROUD-Full stage matrix is reported in Table~\ref{tab:qwen_full_matrix}.

\noindent\textbf{Routed LoRA experts at matched rank.}
Table~\ref{tab:resources}(b) compares separate per-task LoRA experts and PROUD's shared layer at rank $r=32$. PROUD uses the same run as Table~\ref{tab:sensitivity}, with learning rate $2\times10^{-4}$, $\lambda=0.01$, and seed 42, achieving 68.87\% Last and 78.27\% Avg. The experts use their calibrated learning rate $10^{-4}$ and reach 62.91\% Last with CLIP routing~\citep{radford2021learning} and 62.94\% with Oracle routing. PROUD exceeds Oracle routing by 5.94 points in Last and 13.31 in Avg, demonstrating that adapting a shared projector interface with PROUD's subspace optimization outperforms isolated per-task LoRA experts even when routing errors are eliminated.

For inference profiling, we use the original rank-$32$ comparison checkpoints on the same RTX~4090 and 24 fixed UCIT inputs (four per task) for all three modes, with batch size 1, three warm-up generations, and three repetitions. We report mean routing and expert-dispatch time over 72 calls per mode, with all six experts resident on the GPU. Model loading, image I/O, and MLLM input preprocessing are excluded; preprocessing inside the CLIP router is included. Oracle selection uses the true task ID, while merged PROUD has no routing step. Before timing, each mode reproduces the saved predictions for one example per task.

In the initial three-mode profile, we measure greedy generation with exactly 64 output tokens. Including routing, prefill, and decoding, throughput is 42.59, 43.54, and 43.01 tokens/s for CLIP-routed experts, Oracle-routed experts, and PROUD, respectively, computed as total generated tokens divided by summed routing-plus-generation time. Separate one-token generations give median first-token latencies of 127.50, 90.77, and 91.01\,ms, including routing. The same loading, I/O, and preprocessing exclusions apply. These rank-$32$ measurements are distinct from the rank-$128$ merged-inference profiles in Table~\ref{tab:resources}(a).

\begin{table*}[!htbp]
\centering
\proudtabsetup
\caption{Two-seed stability verification (seeds 42 and 43) across ranks and learning rates on UCIT ($\lambda=0.01$).}
\label{tab:app_twoseed}
\begin{tabularx}{\linewidth}{cc*{4}{C{1}}}
\toprule
\rowcolor{proudtableheadbg}
\textbf{Rank $r$} & \textbf{Learning Rate} & \textbf{Seed 42 Last} & \textbf{Seed 43 Last} & \textbf{Last (Mean $\pm$ SD)} $\uparrow$ & \textbf{BWT Mean} $\uparrow$ \\
\midrule
64 & $2\times 10^{-4}$ & 69.74 & 69.11 & $69.42 \pm 0.45$ & -1.20 \\
\rowcolor{proudtableshade}
64 & $1\times 10^{-4}$ & 71.03 & 70.39 & $70.71 \pm 0.45$ & -1.73 \\
128 & $2\times 10^{-4}$ & 71.07 & 68.96 & $70.01 \pm 1.50$ & -1.38 \\
\rowcolor{proudtableshade}
128 & $1\times 10^{-4}$ & 72.71 & 71.63 & $72.17 \pm 0.76$ & -1.10 \\
128 & $5\times 10^{-5}$ & 72.79 & 72.26 & $72.53 \pm 0.38$ & -1.14 \\
\rowcolor{proudtableresult}
\textbf{128} & $\mathbf{8\times 10^{-5}}$ & \textbf{72.49} & \textbf{72.81} & $72.65 \pm 0.23$ & \textbf{-0.82} \\
\bottomrule
\end{tabularx}
\end{table*}

\begingroup
\interlinepenalty=10000
\noindent\textbf{Sensitivity to rank, damping, sample budget, and task order.}
Table~\ref{tab:sensitivity} and Figure~\ref{fig:sensitivity_app} show stable UCIT results across ranks $r\in[32,256]$, damping $\gamma\in\{0.003,0.01,0.03\}$, preview/history budgets, and reversed or random task orders. The rank sweep is evaluated at $\mathrm{lr} = 2 \times 10^{-4}$, with the cross-seed rank and learning-rate grid reported in Table~\ref{tab:app_twoseed}. We retain $\gamma=0.01$ and $\lambda=0.01$ as shared defaults across benchmarks and backbones, rather than using benchmark-specific damping values. This setting preserves strong UCIT retention ($-0.93\%$ BWT), while larger damping ($\gamma=0.03$) raises UCIT Last from 72.49\% to 74.18\% with slightly greater forgetting ($-1.23\%$ BWT).\par
\endgroup

\noindent\textbf{Three-seed replication across backbones.}
Table~\ref{tab:app_multiseed} reports all three completed independent runs for each configuration, with sample standard deviations. On UCIT, the default configuration reaches $72.53\pm0.26\%$ Last with LLaVA and $84.26\pm0.23\%$ with Qwen2.5-VL. All three Qwen runs retain near-zero BWT ($-0.16$ to $+0.17$). Table~\ref{tab:mat_replication_ucit} provides the additional complete stage matrices.

\begin{table*}[!htbp]
\centering
\proudtabsetup
\caption{Three-seed PROUD results on UCIT and CoIN (seeds 42, 43, and 44). The three middle columns show individual Last scores; the final three columns report means $\pm$ sample SD over complete sequences. All runs use $r=128$, preview 256, history 512, and one training epoch per task. Both benchmarks use the default uniform hyperparameters. The main benchmark tables use seed 42.}
\label{tab:app_multiseed}
\footnotesize
\setlength{\tabcolsep}{3pt}
\begin{tabularx}{\linewidth}{lC{0.75}C{0.75}C{0.75}C{1.25}C{1.25}C{1.25}}
\toprule
\rowcolor{proudtableheadbg}
\textbf{Benchmark / Backbone} & \proudtablehead{Seed 42\\Last} & \proudtablehead{Seed 43\\Last} & \proudtablehead{Seed 44\\Last} & \textbf{Last} $\uparrow$ & \textbf{Avg} $\uparrow$ & \textbf{BWT} $\uparrow$ \\
\midrule
\rowcolor{proudtableresult}
UCIT / LLaVA-v1.5 & 72.49 & 72.81 & 72.29 & $72.53\pm0.26$ & $81.80\pm0.27$ & $-0.82\pm0.11$ \\
\rowcolor{proudtableresult}
UCIT / Qwen2.5-VL & 84.17 & 84.53 & 84.09 & $84.26\pm0.23$ & $87.80\pm0.08$ & $-0.02\pm0.17$ \\
\rowcolor{proudtableresult}
CoIN / LLaVA-v1.5 & 60.53 & 60.88 & 59.40 & $60.27\pm0.77$ & $66.37\pm0.37$ & $-3.44\pm0.60$ \\
\bottomrule
\end{tabularx}
\end{table*}

\noindent\textbf{Two-seed rank--LR grid.}
Table~\ref{tab:app_twoseed} reports two-seed verification across ranks $r \in \{64, 128\}$ and learning rates on UCIT. Lowering the learning rate from $2\times 10^{-4}$ to $8\times 10^{-5}$ at $r=128$ improves the two-seed mean Last accuracy from $70.01\%$ to $72.65\%$ while reducing the sample standard deviation by more than $6\times$ and improving mean BWT to $-0.82$.

\begin{table*}[!htbp]
\centering
\proudtabsetup
\caption{Measured multi-task joint-training references on UCIT and CoIN. Joint Decoder-LoRA adapts all 32 decoder layers at rank 32 (79.95M trainable parameters), targeting attention Q/K/V/O and FFN gate/up/down projections under the matched data, exposure budget, and evaluation protocol. Within each benchmark, all three joint-training references share identical data splits, exposure budgets, random seeds, and learning rates ($2 \times 10^{-4}$ on UCIT; $8 \times 10^{-5}$ on CoIN). Percentages in parentheses indicate the ratio of Joint Terminal Layer accuracy to Joint Entire Projector accuracy.}
\label{tab:app_joint_training}
\footnotesize
\setlength{\tabcolsep}{4pt}
\begin{tabularx}{\linewidth}{l*{4}{C{1}}}
\toprule
\rowcolor{proudtableheadbg}
\textbf{Benchmark} & \proudtablehead{\textbf{Continual}\\\textbf{PROUD}} & \proudtablehead{\textbf{Joint Terminal}\\\textbf{Layer (16.8M)}} & \proudtablehead{\textbf{Joint Entire}\\\textbf{Projector (21.0M)}} & \proudtablehead{\textbf{Joint Decoder-LoRA}\\\textbf{(80.0M)}} \\
\midrule
UCIT (6 tasks) & 72.49 & 78.00 (99.2\%) & 78.61 & \textbf{79.14} \\
\rowcolor{proudtableshade}
CoIN (8 tasks) & 60.53 & 69.43 (99.4\%) & 69.83 & \textbf{71.90} \\
\bottomrule
\end{tabularx}
\end{table*}

\noindent\textbf{Joint-training capacity references.}
Table~\ref{tab:app_joint_training} compares continual PROUD against multi-task joint training of the terminal projector layer ($16.78\text{M}$ parameters), the entire two-layer projector ($20.98\text{M}$ parameters), and multi-layer Decoder-LoRA ($79.95\text{M}$ parameters). On both benchmarks, joint training of the terminal layer alone achieves over $99\%$ of the performance of training the entire projector ($78.00\%$ vs.\ $78.61\%$ on UCIT; $69.43\%$ vs.\ $69.83\%$ on CoIN). Relative to rank-32 Joint Decoder-LoRA (79.95M parameters), joint terminal-layer tuning retains 98.6\% (UCIT: 78.00\% vs.\ 79.14\%) and 96.6\% (CoIN: 69.43\% vs.\ 71.90\%) of its accuracy with $4.8\times$ fewer parameters, showing that while decoder adaptation provides additional capacity, the terminal projector layer captures the vast majority of multi-task alignment capacity.

\noindent\textbf{Stationary inputs and the terminal adaptation site.}
Across adaptation sites (Table~\ref{tab:ablations_detailed}), adapting LLM decoder or CLIP vision layers trails the final projector layer (72.49\% Last). Although the first projector layer also receives stationary visual inputs, it trails the final layer by 5.69 points (66.80\% vs.\ 72.49\%). Updates to the first layer pass through the intermediate $\mathrm{GELU}$ nonlinearity and second projection before reaching the LLM, so a quadratic displacement metric at that layer no longer exactly measures displacement of the final visual tokens. Only the terminal projector layer simultaneously enjoys stationary inputs and a direct linear mapping into the LLM token space. This distinction complements the joint-training capacity comparison above: stationarity alone does not explain the difference between the two projector layers.

\begin{table*}[!htbp]
\centering
\proudtabsetup
\caption{Evolution of held-out whitened gradient capture ratio $\operatorname{Capture}_t(G)$ across optimization steps ($0, 30, 150, 300$) and final step-300 validation cross-entropy loss on VizWiz-caption and IconQA. Capture uses the mean loss gradient on a fixed 32-example diagnostic split at the current weights, separate from the 128-example probe evaluation split in Table~\ref{tab:probing}.}
\label{tab:app_trajectory}
\footnotesize
\setlength{\tabcolsep}{2pt}
\begin{tabularx}{\linewidth}{l*{5}{C{1}}*{5}{C{1}}}
\toprule
\rowcolor{proudtableheadbg}
& \multicolumn{5}{c}{\textbf{VizWiz-caption (300 Steps)}} & \multicolumn{5}{c}{\textbf{IconQA (300 Steps)}} \\
\cmidrule(lr){2-6} \cmidrule(lr){7-11}
\rowcolor{proudtableheadbg}
\textbf{Subspace} & \textbf{Step 0} & \textbf{Step 30} & \textbf{Step 150} & \textbf{Step 300} & \textbf{Val CE} $\downarrow$ & \textbf{Step 0} & \textbf{Step 30} & \textbf{Step 150} & \textbf{Step 300} & \textbf{Val CE} $\downarrow$ \\
\midrule
Random & 3.22\% & 3.16\% & 3.20\% & 3.13\% & 2.6584 & 3.62\% & 3.85\% & 3.58\% & 3.55\% & 0.5127 \\
\rowcolor{proudtableshade}
Historical Near-Null & 3.83\% & 3.85\% & 3.94\% & 3.97\% & 3.3139 & 3.44\% & 3.35\% & 4.15\% & 3.99\% & 0.5996 \\
Current Gradient & 9.32\% & 8.13\% & 7.95\% & 7.82\% & 2.4890 & 14.28\% & 11.98\% & 12.86\% & 13.79\% & 0.4787 \\
\midrule
\rowcolor{proudtableresult}
\textbf{PROUD (Ours)} & \textbf{31.57\%} & \textbf{25.87\%} & \textbf{25.11\%} & \textbf{24.74\%} & \textbf{2.1730} & \textbf{58.59\%} & \textbf{72.97\%} & \textbf{40.90\%} & \textbf{47.32\%} & \textbf{0.4636} \\
\bottomrule
\end{tabularx}
\end{table*}

\noindent\textbf{Detailed 300-step gradient capture trajectories.}
Table~\ref{tab:app_trajectory} reports the full step-by-step evolution of $\operatorname{Capture}_t(G)$ across 300 updates. These trajectories use the mean gradient on a fixed 32-example held-out diagnostic split, whereas the single-step probes in Table~\ref{tab:probing} and Figure~\ref{fig:teaser}(b) use the mean gradient on a separate 128-example evaluation split. Both use the same capture definition, but the diagnostic gradients differ; consequently, the step-0 values (31.57\% and 58.59\%) need not equal the single-step probe values (39.77\% and 86.48\%). Throughout training, the fixed input basis $P_t$ selected by PROUD captures $6\times$ to $12\times$ more whitened gradient energy than Random Basis and Historical Near-Null subspaces, and $2\times$ to $3.5\times$ more than the Current Gradient subspace.


\begin{table*}[!htbp]
\centering
\proudtabsetup
\caption{CoIN Stage-7 regularization sweep with a uniform learning rate of $8\times10^{-5}$ (LLaVA-v1.5-7B, $r=128$, one epoch). Only the VQAv2 penalty changes; all other stages use $\lambda=0.01$.}
\label{tab:app_coin_lambda}
\footnotesize
\setlength{\tabcolsep}{2pt}
\begin{tabularx}{\linewidth}{>{\raggedright\arraybackslash}p{0.23\linewidth}C{0.95}C{1.20}C{1.00}C{0.75}C{1.00}C{1.00}C{1.00}C{1.20}C{1.00}C{0.90}C{1.00}}
\toprule
\rowcolor{proudtableheadbg}
\textbf{CoIN Configuration} & \textbf{SciQA} & \textbf{TextVQA} & \textbf{ImgNet} & \textbf{GQA} & \textbf{VizWiz} & \textbf{Ground.} & \textbf{VQAv2} & \textbf{OCRVQA} & \proudtablehead{\textbf{Last} $\uparrow$} & \proudtablehead{\textbf{Avg} $\uparrow$} & \proudtablehead{\textbf{BWT} $\uparrow$} \\
\midrule
Uniform default ($\lambda=0.01$ all stages) & 75.67 & \textbf{56.18} & 66.46 & \textbf{57.36} & \textbf{52.56} & 51.92 & \textbf{62.68} & \textbf{61.41} & 60.53 & 66.59 & -3.66 \\
\rowcolor{proudtableshade}
VQAv2 $\lambda=0.1$ & 75.88 & 55.92 & 74.55 & 56.90 & 51.63 & 51.88 & 60.89 & 61.29 & 61.12 & 66.75 & -2.72 \\
\rowcolor{proudtableresult}
\textbf{VQAv2 $\lambda=0.3$} & \textbf{75.90} & 55.85 & \textbf{77.49} & 56.72 & 51.24 & \textbf{52.30} & 60.20 & 61.20 & \textbf{61.36} & \textbf{66.81} & \textbf{-2.33} \\
\bottomrule
\end{tabularx}
\end{table*}

\noindent\textbf{Stage-specific regularization sensitivity and task-granularity analysis on CoIN.}
During Stage~7 (VQAv2), none of five earlier tasks (ScienceQA, TextVQA, GQA, VizWiz, and Grounding) loses more than 1.47 percentage points; TextVQA and VizWiz improve by 0.76 and 2.22 points, respectively (Table~\ref{tab:mat_coin_llava}(a)). ImageNet instead drops from 86.53\% to 67.15\%. Two factors may contribute to this localized decline. First, \emph{exposure imbalance}: VQAv2 contributes $82{,}752$ recorded sample exposures, $1.53\times$ the prior-stage mean of $54{,}256$, increasing update steps under a uniform learning rate and $\lambda=0.01$. Second, a possible \emph{label-granularity shift on shared visual features}: intensive open-ended QA supervision may favor coarse natural-object concepts over ImageNet's fine-grained class distinctions. ImageNet contributes only one sixth of the stage-balanced metric $C_{<7}=\frac{1}{6}\sum_{s<7}\widehat{C}_s$, which controls average token displacement rather than individual classification boundaries. To isolate the Stage-7 penalty, Table~\ref{tab:app_coin_lambda} holds every stage's learning rate at $8\times10^{-5}$ and all other penalties at $0.01$. Increasing only $\lambda_7$ from $0.01$ to $0.1$ and $0.3$ raises final ImageNet retention from $66.46\%$ to $74.55\%$ and \textbf{77.49\%}, Last from $60.53\%$ to $61.12\%$ and \textbf{61.36\%}, and BWT from $-3.66\%$ to $-2.72\%$ and \textbf{$-2.33\%$}. These controls confirm the benefit of stronger magnitude regularization at high-exposure stages.

\FloatBarrier
\FloatBarrier
\subsection{Regularization and cumulative interface drift}
\label{app:regularization_diagnostics}

We measure cumulative displacement relative to the projector weights immediately after learning each earlier task. For task $s$ after stage $t>s$, define
\begin{equation}
    D_{s,t}=\operatorname{tr}\!\left((W_t-W_s)\widehat{C}_s(W_t-W_s)^\top\right),
    \qquad \overline{D}_t=\frac{1}{t-1}\sum_{s=1}^{t-1}D_{s,t}.
    \label{eq:observed_cumulative_drift}
\end{equation}
We compute these quantities in FP64 from saved FP32 weights and recover each $\widehat{C}_s$ from the stored stage-balanced running averages. Thus, the diagnostic measures displacement under the retained empirical training statistics. Each earlier task has its own reference $W_s$; the final average on UCIT covers the first five tasks.

\begin{figure}[htbp]
\centering
\includegraphics[width=\linewidth]{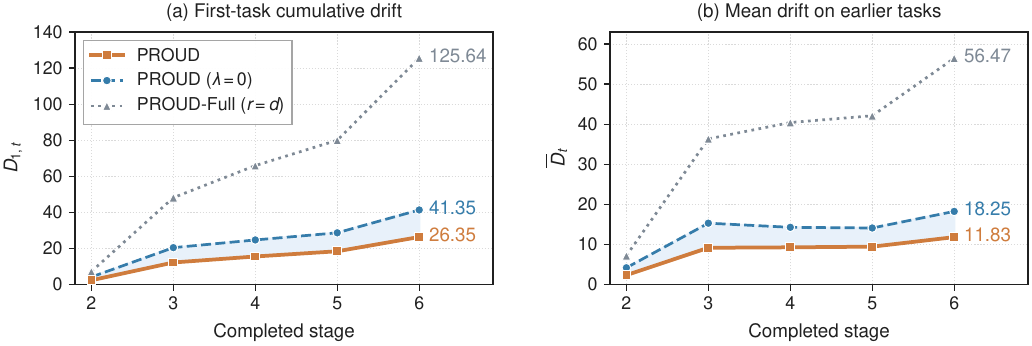}
\caption{\textbf{Cumulative interface displacement on UCIT.} (a) Displacement on the first task, anchored at $W_1$. (b) Mean displacement on all earlier tasks, each anchored at its own post-training weight $W_s$. The set averaged in (b) grows with the stage index. All points are computed from saved checkpoints and retained moments.}
\label{fig:cumulative_drift}
\end{figure}

Table~\ref{tab:cumulative_drift_results} compares the default PROUD configuration ($\lambda=0.01$) against removing the explicit Frobenius penalty ($\lambda=0$) and PROUD-Full. The metric-whitened rank-$r$ basis $P_t=L_t^{-\top}V_{t,r}$ preconditions update directions via $L_t^{-\top}$ and yields cumulative interface displacement $\bar{D}_6=18.25$, compared with $56.47$ for PROUD-Full. Adding the explicit isometric penalty ($\lambda=0.01$) further reduces $\bar{D}_6$ by 35.2\% ($18.25\to11.83$; Figure~\ref{fig:cumulative_drift}) and first-task drift $D_{1,6}$ by 36.3\% ($41.35\to26.35$). Meanwhile, Table~\ref{tab:app_coin_lambda} shows how increasing $\lambda$ improves task retention under high-exposure training.

\begin{table}[htbp]
\centering
\proudtabsetup
\caption{Cumulative interface displacement and isometry diagnostics after six UCIT stages (seed 42). $\overline{D}_6$ averages the first five tasks; $D_{1,6}$ measures first-task displacement. Bold marks the lowest displacement. Isometry error is the maximum stage-end relative error across six stages. All runs use the terminal projector layer and one-pass training; full-space adaptation has a larger rank budget.}
\label{tab:cumulative_drift_results}
\begin{tabularx}{\linewidth}{>{\raggedright\arraybackslash}p{0.29\linewidth}*{4}{C{1}}}
\toprule
\rowcolor{proudtableheadbg}
\textbf{Method} & \textbf{Rank} & $\overline{D}_6$ $\downarrow$ & $D_{1,6}$ $\downarrow$ & \proudtablehead{Isometry error\\($\times10^{-8}$)} \\
\midrule
PROUD-Full ($r=d$) & 4096 & 56.47 & 125.64 & 10.75 \\
PROUD ($\lambda=0$) & 128 & 18.25 & 41.35 & 19.03 \\
\rowcolor{proudtableresult}
PROUD ($\lambda=0.01$) & 128 & \textbf{11.83} & \textbf{26.35} & 6.12 \\
\bottomrule
\end{tabularx}
\end{table}

\noindent\textbf{Historical cost and the isometric identity.}
At each stage end, we separate $q_t(\Delta W_t)$ into its empirical historical component and the isotropic damping term from Eq.~\eqref{eq:drift_decomposition}. Table~\ref{tab:observed_cost_diagnostics} reports their values and the observed relative error in Eq.~\eqref{eq:isometry}. For Stages 2--6, the historical component accounts for $75.7\%$--$84.1\%$ of the total cost. Across all six stages, the relative isometry error is below $6.2\times10^{-8}$. Stage 1 has no historical term because $C_{<1}=0$.

\begin{table}[htbp]
\centering
\proudtabsetup
\caption{Observed stage-end cost decomposition and isometry diagnostics for default PROUD on UCIT (seed 42). Historical share is the historical component divided by the sum of the two cost columns.}
\label{tab:observed_cost_diagnostics}
\begin{tabularx}{\linewidth}{c*{4}{C{1}}}
\toprule
\rowcolor{proudtableheadbg}
\textbf{Stage} & \proudtablehead{Historical\\component} & \proudtablehead{Damping\\component} & \proudtablehead{Historical\\share (\%)} & \proudtablehead{Relative isometry error\\($\times10^{-8}$)} \\
\midrule
1 & 0.0000 & 1.6437 & 0.00 & 1.56 \\
\rowcolor{proudtableshade}
2 & 2.3505 & 0.7559 & 75.67 & 3.05 \\
3 & 7.9956 & 1.7405 & 82.12 & 6.12 \\
\rowcolor{proudtableshade}
4 & 3.1466 & 0.5931 & 84.14 & 3.09 \\
5 & 2.4847 & 0.5560 & 81.72 & 5.85 \\
\rowcolor{proudtableshade}
6 & 4.0951 & 0.8184 & 83.34 & 2.56 \\
\bottomrule
\end{tabularx}
\end{table}

\FloatBarrier
\subsection{Training and preview budgets}
\label{app:budget_studies}

\begin{figure}[htbp]
\centering
\includegraphics[width=\linewidth]{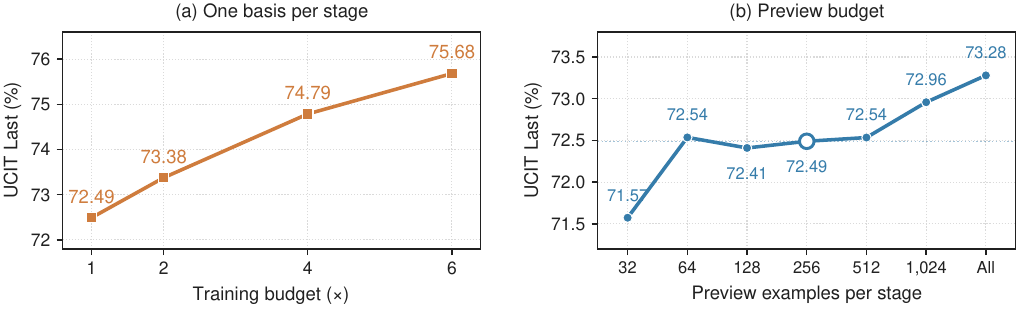}
\caption{\textbf{Training and preview budgets on UCIT.} (a) Last accuracy improves with additional training while each stage retains its initially selected basis. (b) Preview-budget comparison with one-pass training and 512 historical-statistics samples per task. Preview sizes are ordered categories; the open marker denotes the 256-example default, and All uses the full current-task dataset. Both studies use seed 42.}
\label{fig:training_preview_budgets}
\end{figure}

\noindent\textbf{Longer training with one basis per stage.}
We keep $r=128$, learning rate $8\times10^{-5}$, $\lambda=0.01$, and seed 42 fixed while increasing the training exposure budget to $1\times$, $2\times$, $4\times$, and $6\times$. Each stage still selects its input basis only once and updates the historical statistics once after training; the learning-rate schedule spans the full training budget of that stage. Last improves from $72.49\%$ to $73.38\%$, $74.79\%$, and $75.68\%$ (Figure~\ref{fig:training_preview_budgets}(a)). These results show that further learning remains possible within the initially selected subspace. Table~\ref{tab:training_budget} reports exposure counts; main comparisons use $1\times$.

\begin{table}[htbp]
\centering
\proudtabsetup
\caption{Training-budget comparison on UCIT with a fixed input basis throughout each stage ($r=128$, $\lambda=0.01$, seed 42). Exposures count training examples processed, including repetitions. Best scores are in bold. All four runs select one basis per task; their training budgets differ.}
\label{tab:training_budget}
\begin{tabularx}{\linewidth}{C{0.8}C{1.5}C{0.9}C{0.9}C{0.9}}
\toprule
\rowcolor{proudtableheadbg}
\textbf{Budget} & \textbf{Training exposures} & \textbf{Last} $\uparrow$ & \textbf{Avg} $\uparrow$ & \textbf{BWT} $\uparrow$ \\
\midrule
\rowcolor{proudtableresult}
$1\times$ & 213,792 & 72.49 & 81.89 & \textbf{-0.93} \\
$2\times$ & 427,584 & 73.38 & 83.11 & -2.18 \\
\rowcolor{proudtableshade}
$4\times$ & 855,168 & 74.79 & \textbf{83.72} & -1.90 \\
$6\times$ & 1,282,752 & \textbf{75.68} & 83.56 & -1.07 \\
\bottomrule
\end{tabularx}
\end{table}

\noindent\textbf{Preview sample budget.}
We compare 32, 64, 128, 256, 512, 1,024, and all available preview examples per task, keeping one-pass optimization and $n_s=512$ historical-statistics samples fixed. The remaining settings are $r=128$, learning rate $8\times10^{-5}$, $\lambda=0.01$, and seed 42. Using 64 or 128 examples reaches 72.54\% or 72.41\% Last, respectively, within 0.08 points of the 256-example default while using one quarter or one half of its preview samples. Reducing the preview to 32 examples gives 71.57\%. The full-preview run reaches $73.28\%$ Last, $0.79$ points above the default (Table~\ref{tab:preview_budget}, Figure~\ref{fig:training_preview_budgets}(b)); its stage preparation totals $15{,}124.30\text{ s}$ ($4.20\text{ h}$) across six tasks. The first-stage preview also calibrates $\rho$ under the original procedure, so this comparison varies both the gradient estimate and the damping estimate.

\begin{table}[htbp]
\centering
\proudtabsetup
\caption{Preview-budget comparison on UCIT (seed 42). Optimization uses the same 213,792 training exposures and 512 historical-statistics samples per task. Full preview uses 213,857 available examples across the six tasks. Best scores are in bold.}
\label{tab:preview_budget}
\begin{tabularx}{\linewidth}{C{1.6}C{0.8}C{0.8}C{0.8}}
\toprule
\rowcolor{proudtableheadbg}
\textbf{Preview examples per task} & \textbf{Last} $\uparrow$ & \textbf{Avg} $\uparrow$ & \textbf{BWT} $\uparrow$ \\
\midrule
32 & 71.57 & 80.79 & -0.47 \\
\rowcolor{proudtableshade}
64 & 72.54 & 81.42 & -0.22 \\
128 & 72.41 & 81.63 & -0.64 \\
\rowcolor{proudtableresult}
256 & 72.49 & 81.89 & -0.93 \\
512 & 72.54 & 81.43 & -0.62 \\
\rowcolor{proudtableshade}
1,024 & 72.96 & \textbf{81.93} & -0.79 \\
All & \textbf{73.28} & 81.89 & \textbf{-0.21} \\
\bottomrule
\end{tabularx}
\end{table}

\FloatBarrier
\subsection{Test-instruction variation and output-sensitivity control}
\label{app:additional_controls}

\noindent\textbf{Test-instruction variation.}
We evaluate each fixed final UCIT checkpoint with two alternative instruction templates, without retraining. Both variants paraphrase only the answer-style suffix, preserving the images, questions, options, rationales, answers, and example order. For instance, the caption instruction becomes ``Write a brief caption describing the image'' in Template 1 and ``Provide a short caption for this image'' in Template 2. IconQA fill-in examples without a suffix instead receive a neutral instruction prefix. Table~\ref{tab:template_robustness} reports all four models under the original and both modified templates. PROUD obtains 72.64\% and 71.67\% Last, within 0.83 points of its original 72.49\%, and retains its advantage over sequential LoRA for both tested variations. PROUD-Full remains close, at 71.86\% and 71.11\%.

\begin{table}[htbp]
\centering
\proudtabsetup
\caption{UCIT Last accuracy under two test-instruction paraphrases. The same final checkpoints and scoring protocol are used in all three columns. Best scores are in bold. Each template evaluates the same 3,000 test examples per task across all six tasks.}
\label{tab:template_robustness}
\begin{tabularx}{\linewidth}{>{\raggedright\arraybackslash}Xrrr}
\toprule
\rowcolor{proudtableheadbg}
\textbf{Method} & \textbf{Original} & \textbf{Template 1} & \textbf{Template 2} \\
\midrule
Zero-shot & 33.04 & 31.31 & 36.66 \\
\rowcolor{proudtableshade}
Seq.\ LoRA & 47.63 & 47.96 & 47.53 \\
PROUD-Full ($r=d$) & 72.28 & 71.86 & 71.11 \\
\rowcolor{proudtableresult}
\textbf{PROUD} & \textbf{72.49} & \textbf{72.64} & \textbf{71.67} \\
\bottomrule
\end{tabularx}
\end{table}

\newpage
\noindent\textbf{Adding output sensitivity to the historical metric.}
We compare PROUD's input second-moment metric with a factorized input--output sensitivity variant. At each stage end, the variant additionally collects loss-gradient second moments at the projector output from 512 current-task examples (up to 32 visual tokens per image). Each output moment is normalized to unit mean eigenvalue and averaged equally across stages. After isotropic damping, the resulting output metric $S_t$ modifies subspace selection through $\bar{G}_t^\top S_t^{-1}\bar{G}_t$ and replaces the in-stage penalty by $\frac{\lambda}{2}\operatorname{tr}(B_t^\top S_t B_t)$. This is a factorized empirical sensitivity approximation. Both methods use the same terminal layer, rank, learning rate, training exposure, preview budget, and input-statistics budget.

Table~\ref{tab:output_sensitivity} shows that the input-only metric reaches 72.49\% Last, compared with 72.38\% for the output-sensitive variant. The latter has slightly better BWT ($-0.75$ vs.\ $-0.93$) and requires additional output-gradient statistics, whose collection takes 201.93\,s across six stages. Thus, the simpler input-only metric achieves comparable continual performance without requiring additional output-gradient collection.

\begin{table}[htbp]
\centering
\proudtabsetup
\caption{Input-only versus factorized input--output sensitivity metrics on UCIT ($r=128$, seed 42). Best scores are in bold. Both methods train 0.52M parameters. The output-sensitive variant collects 512 additional output-gradient samples per stage (3,072 in total). Its 201.93\,s collection time measures that added substep, not total setup or training time.}
\label{tab:output_sensitivity}
\begin{tabularx}{\linewidth}{>{\raggedright\arraybackslash}Xrrr}
\toprule
\rowcolor{proudtableheadbg}
\textbf{Historical metric} & \textbf{Last} $\uparrow$ & \textbf{Avg} $\uparrow$ & \textbf{BWT} $\uparrow$ \\
\midrule
Input--output sensitivity & 72.38 & 81.65 & \textbf{-0.75} \\
\rowcolor{proudtableresult}
\textbf{PROUD (Input only)} & \textbf{72.49} & \textbf{81.89} & -0.93 \\
\bottomrule
\end{tabularx}
\end{table}

\FloatBarrier
\subsection{Same-layer controls and segmented stage-boundary stress tests}
\label{app:projector_controls}

\noindent\textbf{Same-layer InfLoRA/LoRA-GA and parameter-matched LoRA controls.}
Table~\ref{tab:same_layer_controls} provides two additional control groups at the terminal projector layer under the exact same one-epoch UCIT training budget. First, direct ports of InfLoRA~\citep{liang2024inflora} and LoRA-GA~\citep{wang2024loraga} at rank $r = 128$---which lack our metric-orthonormal normalization ($P_Q^\top M_t P_Q = I_r$) and in-stage isometric drift regularization in Eq.~\eqref{eq:stage_objective}---reach only $54.94\%$ and $55.84\%$ Last. Equipping their respective subspace criteria with our metric normalization and isometric penalty in Table~\ref{tab:main_cl_detailed} substantially improves them to $61.01\%$ and $67.89\%$, yet both still trail PROUD ($72.49\%$). Second, under an exact 524,288-parameter budget, rank-64 LoRA~\citep{hu2022lora} (47.36\%) and rank-64 LoRA with isotropic weight decay (47.17\%) both suffer severe forgetting ($-22.77\%$ to $-23.01\%$ BWT). Adding our historical drift penalty improves rank-64 LoRA to 60.46\% Last ($-6.43\%$ BWT) yet still trails PROUD by 12.03 points, while freezing a random input factor (LoRA-FA style~\citep{zhang2023lorafa}, $r=128$) reaches 49.41\% Last. These controls confirm that PROUD's gains do not stem merely from terminal-layer placement, parameter count, isotropic weight decay, or freezing one factor, but from coupling metric-whitened subspace selection with isometric regularization.

\begin{table}[!htbp]
\centering
\proudtabsetup
\caption{Same-layer adaptations and strictly parameter-matched LoRA controls on UCIT with LLaVA-v1.5-7B (seed 42). InfLoRA trains 524,288 parameters, while LoRA-GA trains both factors, totaling 1,048,576 parameters. Every method in the second group, including PROUD, trains exactly 524,288 parameters.}
\label{tab:same_layer_controls}
\begin{tabularx}{\linewidth}{>{\raggedright\arraybackslash}Xrrrrr}
\toprule
\rowcolor{proudtableheadbg}
\textbf{Method / Control (Terminal Projector)} & \textbf{Params} & \textbf{Rank} & \textbf{Last} $\uparrow$ & \textbf{Avg} $\uparrow$ & \textbf{BWT} $\uparrow$ \\
\midrule
\multicolumn{6}{l}{\emph{Generative-MCIT adaptations at the terminal projector}} \\
InfLoRA (projector-adapted)~\citep{liang2024inflora} & 0.52M & 128 & 54.94 & 58.01 & -11.87 \\
\rowcolor{proudtableshade}
LoRA-GA (projector-adapted)~\citep{wang2024loraga} & 1.05M & 128 & 55.84 & 70.01 & -23.20 \\
\midrule
\multicolumn{6}{l}{\emph{Strictly parameter-matched LoRA controls (524,288 trainable parameters)}} \\
LoRA ($r=64$)~\citep{hu2022lora} & 0.52M & 64 & 47.36 & 54.13 & -22.77 \\
\rowcolor{proudtableshade}
LoRA ($r=64$) + Weight Decay & 0.52M & 64 & 47.17 & 54.42 & -23.01 \\
LoRA ($r=64$) + Historical Drift Reg & 0.52M & 64 & 60.46 & 67.51 & -6.43 \\
\rowcolor{proudtableshade}
Fixed Random $A$ (LoRA-FA style)~\citep{zhang2023lorafa} & 0.52M & 128 & 49.41 & 49.31 & -6.57 \\
\rowcolor{proudtableresult}
\textbf{PROUD (Ours)} & \textbf{0.52M} & \textbf{128} & \textbf{72.49} & \textbf{81.89} & \textbf{-0.93} \\
\bottomrule
\end{tabularx}
\end{table}

\noindent\textbf{Multi-boundary segmented stress test.}
To test stability under denser stage boundaries, Table~\ref{tab:segmented_stress} splits each of the six UCIT tasks into 2 or 3 sequential chunks (12 and 18 sequential stages, respectively) while preserving the same six domains and total training exposures. Each boundary triggers an optimizer reset, new basis extraction, weight merge, and second-moment update. By holding total data exposure fixed, this protocol directly isolates the effect of stage-boundary accumulation across repeated basis extractions and weight merges: final accuracy stays closely aligned across 6, 12, and 18 stages ($72.49\%$, $72.74\%$, and $71.83\%$ Last), while BWT changes modestly from $-0.93\%$ to $-1.51\%$ and $-2.36\%$.

\begin{table}[!htbp]
\centering
\proudtabsetup
\caption{UCIT segmentation stress test with the same six domains and 213,792 total training exposures (seed 42). Last and BWT are computed at the six original task-completion boundaries, using the definitions in Section~\ref{app:metrics}.}
\label{tab:segmented_stress}
\begin{tabularx}{\linewidth}{>{\raggedright\arraybackslash}Xrrrr}
\toprule
\rowcolor{proudtableheadbg}
\textbf{UCIT Segmentation Protocol} & \textbf{Sequential Stages} & \textbf{Segments / Task} & \textbf{Last} $\uparrow$ & \textbf{BWT} $\uparrow$ \\
\midrule
Standard UCIT & 6 & 1 & 72.49 & -0.93 \\
\rowcolor{proudtableshade}
2-way Segmented Stress Test & 12 & 2 & 72.74 & -1.51 \\
3-way Segmented Stress Test & 18 & 3 & 71.83 & -2.36 \\
\bottomrule
\end{tabularx}
\end{table}

\begin{table}[!htbp]
\centering
\proudtabsetup
\caption{Complete six-stage UCIT matrix for PROUD-Full on Qwen2.5-VL-7B ($r=d=5120$, 18.35M trainable parameters, seed 42). The default rank-128 PROUD matrix is in Table~\ref{tab:mat_ucit_qwen}. Both use the same task order, one-epoch exposure budget, learning rate $8\times10^{-5}$, and $\lambda=\gamma=0.01$.}
\label{tab:qwen_full_matrix}
\footnotesize
\setlength{\tabcolsep}{4pt}
\begin{tabularx}{\linewidth}{l*{7}{C{1}}}
\toprule
\rowcolor{proudtableheadbg}
\textbf{Stage} & \textbf{ImgNet-R} & \textbf{ArxivQA} & \textbf{VizWiz} & \textbf{IconQA} & \textbf{CLEVR} & \textbf{Flickr30k} & \textbf{Seen Avg} \\
\midrule
T1 & \textbf{90.30} & -- & -- & -- & -- & -- & 90.30 \\
\rowcolor{proudtableshade}
T2 & 90.27 & \textbf{95.33} & -- & -- & -- & -- & 92.80 \\
T3 & 90.40 & 95.33 & \textbf{62.78} & -- & -- & -- & 82.84 \\
\rowcolor{proudtableshade}
T4 & 90.27 & 95.43 & 63.28 & \textbf{94.37} & -- & -- & 85.84 \\
T5 & 90.33 & 95.47 & 63.02 & 94.60 & \textbf{98.83} & -- & 88.45 \\
\rowcolor{proudtableshade}
T6 & 90.07 & 95.43 & 63.44 & 94.67 & 98.87 & \textbf{54.62} & \textbf{82.85} \\
\midrule
\multicolumn{8}{c}{$\text{Last}=82.85\%,\quad \text{Avg}=87.18\%,\quad \text{BWT}=+0.17\%$} \\
\bottomrule
\end{tabularx}
\end{table}

\FloatBarrier
\section{Detailed stage-onset decomposition timing}
\label{app:efficiency}

Table~\ref{tab:resources}(a) in the main paper reports training resources and RTX~4090 inference profiles on UCIT with $m=d=4096$ and $r=128$. Adapter and Adam-state memory are measured in FP32. All compared methods have the same peak inference memory of 14.10\,GiB after weight merging.

\noindent\textbf{Persistent memory and merged inference.}
Retaining $C_{<t}$ in FP32 requires a fixed 64.0\,MiB buffer for $d=4096$; symmetric storage would reduce this to approximately 32\,MiB. This historical-statistics buffer is separate from the adapter and Adam-state memory in Table~\ref{tab:resources}. Freezing $P_t$ halves Adam states relative to training both low-rank factors (4.0 vs.\ 8.0\,MiB). After each stage, the update is merged into $W_t$, giving zero inference parameter growth or routing overhead; the RTX~4090 profiles in Table~\ref{tab:resources} report the measured latency and memory footprint.

Table~\ref{tab:app_timing} reports stage-onset setup and matrix-decomposition time across the six UCIT stages ($d=m=4096$), profiled independently on CPU in FP64 with four threads per process and the maximum time across four workers. Projector training and weight merging use FP32, whereas decompositions and residual checks in Section~\ref{app:numerical_checks} use FP64. Cholesky factorization of $M_t$ takes less than $0.17\text{ s}$ per stage, and SVD of $K_t$ takes $\sim21.7\text{ s}$ per stage. Total matrix decomposition requires $132.64\text{ s}$ across all six stages ($0.75\%$ of recorded training time), and total stage-onset preparation including preview gradients and feature-statistics collection requires $272.89\text{ s}$ ($1.55\%$ of recorded training time).

\begin{table}[h]
\centering
\proudtabsetup
\caption{Breakdown of stage-onset setup and decomposition time across the six UCIT stages ($d=m=4096$). Setup was profiled separately across the six saved stage checkpoints. Percentages are estimates relative to the recorded FP32 training duration of $17{,}624.86\text{ s}$.}
\label{tab:app_timing}
\begin{tabularx}{\linewidth}{l*{3}{C{1}}}
\toprule
\rowcolor{proudtableheadbg}
\textbf{Stage-Onset Step} & \textbf{Per-Stage Range (s)} & \textbf{6-Stage Total (s)} & \textbf{Share of Training Time} \\
\midrule
Cholesky factorization ($M_t = L_t L_t^\top$) & 0.159 -- 0.168 & 0.98 & 0.006\% \\
\rowcolor{proudtableshade}
Whitened SVD ($K_t = U_t \Sigma_t V_t^\top$) & 21.56 -- 21.80 & 130.02 & 0.738\% \\
Total matrix decomposition (incl.\ basis solve) & 21.99 -- 22.23 & 132.64 & 0.753\% \\
\rowcolor{proudtableshade}
Preview gradient \& feature statistics collection & 19.73 -- 29.25 & 140.26 & 0.796\% \\
\midrule
\rowcolor{proudtableresult}
\textbf{Total stage-onset setup overhead} & \textbf{41.72 -- 51.48} & \textbf{272.89} & \textbf{1.548\%} \\
\bottomrule
\end{tabularx}
\end{table}

\FloatBarrier
\section{Numerical verification checklist}
\label{app:numerical_checks}

Table~\ref{tab:app_numerical} summarizes the runtime checks and supporting unit tests for the identities in Section~\ref{sec:methodology}. Second-moment accumulation, projector training, and weight merging use FP32. Cholesky factorization, SVD, and triangular solves are executed on CPU in FP64, without explicit matrix inversion or forming $\bar{G}_t^\top\bar{G}_t$. The resulting basis $P_t$ is cast to FP32 for training and merging. Metric-orthogonality and cost-identity residuals are evaluated in FP64 from the stored FP32 tensors.

For the default six-stage UCIT run, the observed relative error in the isometric cost identity ranges from $1.56\times10^{-8}$ to $6.12\times10^{-8}$, well below the $5\times10^{-4}$ acceptance threshold. Stage-specific measurements and the empirical historical cost fractions are reported in Table~\ref{tab:observed_cost_diagnostics}.

\begin{table*}[!htbp]
\centering
\proudtabsetup
\caption{Numerical checks aligned with the implementation. Decompositions use CPU FP64; training tensors and weight merges use FP32. Runtime gates and supporting unit-test comparisons are distinguished below.}
\label{tab:app_numerical}
\footnotesize
\setlength{\tabcolsep}{3pt}
\begin{tabularx}{\linewidth}{>{\raggedright\arraybackslash}p{0.18\linewidth}>{\raggedright\arraybackslash}p{0.57\linewidth}>{\raggedright\arraybackslash}X}
\toprule
\rowcolor{proudtableheadbg}
\textbf{Verification Item} & \textbf{Mathematical Identity or Error Audited} & \textbf{Acceptance Criterion} \\
\midrule
1.\ Zero-Init Invariance & $B_t^{(0)}=\mathbf{0}$ and $f(I;W_{t-1}+B_t^{(0)}P_t^\top)=f(I;W_{t-1})$ & Zero nonzero entries in $B_t^{(0)}$; exact output match in unit tests \\
\rowcolor{proudtableshade}
2.\ Metric Orthonormality & $\max_{i,j}|(P_t^\top M_t P_t-I_r)_{ij}|$, evaluated after storing $P_t$ in FP32 & Finite and $\le 5\times10^{-4}$ \\
3.\ Isometric Cost Identity & $\big|q_t(B_tP_t^\top)-\|B_t\|_F^2\big|/\max(\|B_t\|_F^2,10^{-30})$ & Relative error $\le 5\times10^{-4}$ \\
\rowcolor{proudtableshade}
4.\ Empirical Drift Match & $\mathbb{E}_{\hat{\mathcal H}}\|\Delta W h\|_2^2+\rho\|\Delta W\|_F^2=\|B\|_F^2$ for $\Delta W=BP^\top$, $P^\top M P=I$ & Verified within FP32 numerical tolerance ($\le 10^{-5}$) \\
5.\ Subspace Grad Projection & $\nabla_B(\mathcal L+\tfrac{\lambda}{2}\|B\|_F^2)=(\nabla_W\mathcal L)P+\lambda B$ & Unit-test absolute/relative tolerances $10^{-7}/10^{-4}$ \\
\rowcolor{proudtableshade}
6.\ Lossless Stage Merging & $\|z_{\mathrm{merged}}-z_{\mathrm{unmerged}}\|_F/\max(\|z_{\mathrm{unmerged}}\|_F,10^{-12})$ on the FP32 probe & Finite and $\le 5\times10^{-5}$ \\
\bottomrule
\end{tabularx}
\end{table*}

\FloatBarrier
\section*{LLM Usage Statement}

ChatGPT (OpenAI) and Gemini (Google) were used in an assistive capacity to provide reviewer-style feedback on manuscript drafts and suggestions for language refinement, related work and citations, code implementation, experimental design and analysis, figure presentation, and research framing. All AI-assisted suggestions were reviewed and evaluated by the authors before incorporation. No generative AI tools were used for data generation, dataset creation, or annotation. The authors take full responsibility for the technical content, experimental results, and scientific claims presented in this paper.

\end{document}